\documentclass[journal]{IEEEtai}

\usepackage[colorlinks,urlcolor=blue,linkcolor=blue,citecolor=blue]{hyperref}

\usepackage{color,array}

\usepackage{graphicx}

\usepackage{amsmath,amssymb}
\usepackage{bm}

\usepackage{booktabs}
\usepackage{multirow}
\usepackage{tabularx}

\begin{document}

\title{Deviation-Aware Pretraining for Cardiovascular Disease Detection from Electrocardiograms}

% \author{First A. Author, \IEEEmembership{Fellow, IEEE}, Second B. Author, and Third C. Author, Jr., \IEEEmembership{Member, IEEE}
% \thanks{This paragraph of the first footnote will contain the date on which you submitted your paper for review. It will also contain support information, including sponsor and financial support acknowledgment. For example, ``This work was supported in part by the U.S. Department of Commerce under Grant BS123456.'' }
% \thanks{The next few paragraphs should contain the authors' current affiliations, including current address and e-mail. For example, F. A. Author is with the National Institute of Standards and Technology, Boulder, CO 80305 USA (e-mail: author@boulder.nist.gov).}
% \thanks{S. B. Author, Jr., was with Rice University, Houston, TX 77005 USA. He is now with the Department of Physics, Colorado State University, Fort Collins, CO 80523 USA (e-mail: author@lamar.colostate.edu).}
% \thanks{T. C. Author is with the Electrical Engineering Department, University of Colorado, Boulder, CO 80309 USA, on leave from the National Research Institute for Metals, Tsukuba, Japan (e-mail: author@nrim.go.jp).}
% \thanks{This paragraph will include the Associate Editor who handled your paper.}}
\author{Siyuan Yan, Haibo Zheng, Feng Lu, Wei Li}

% \markboth{Journal of IEEE Transactions on Artificial Intelligence, Vol. 00, No. 0, Month 2020}
% {First A. Author \MakeLowercase{\textit{et al.}}: Bare Demo of IEEEtai.cls for IEEE Journals of IEEE Transactions on Artificial Intelligence}

\maketitle

\begin{abstract}
Electrocardiogram (ECG) signals are critical for cardiovascular disease (CVD) detection. Most current deep learning models rely on supervised, task-specific training, which limits their ability to generalize across diseases, patient populations, and acquisition settings. This paper presents a self-supervised ECG feature extraction framework that learns from nearly one million unlabeled 12-lead ECG recordings. The framework includes two modules: Deviation Localization Embedding (DLE), which detects waveform-level deviations from normal patterns, and Association Reconstruction Embedding (ARE), which models cross-lead associations through deviation-guided reconstruction to capture abnormal patterns expressed across multiple leads.  These modules enable the framework to capture subtle deviation-related patterns and their cross-lead context. The framework achieves up to 6.5\% AUROC and 12.8\% F1 improvements over prior methods on multiple public datasets, even without additional fine-tuning the pretrained feature extractor. Embedding visualizations show that the learned features support meaningful disease grouping and anomaly identification. The approach reduces the need for labeled data and can support clinicians in diverse healthcare settings. The framework has potential to improve early detection and monitoring of cardiovascular diseases, contributing to more accessible and equitable heart care.
\end{abstract}

\begin{IEEEImpStatement}
This work provides a self-supervised ECG feature extraction framework that can learn transferable representations from large-scale unlabeled 12-lead ECG recordings. By reducing dependence on disease-specific labels and downstream fine-tuning, the framework may lower the cost of developing ECG analysis models for diverse clinical scenarios. The proposed deviation localization and association reconstruction modules help identify subtle waveform abnormalities and cross-lead pathological patterns, offering more interpretable features for cardiovascular disease detection. Its generalization across multiple public and clinical datasets suggests potential value for early screening, risk assessment, and long-term monitoring, especially in settings where expert annotations are limited. More broadly, this study supports the development of reusable ECG representation models that may improve the accessibility, scalability, and reliability of AI-assisted cardiovascular care.
\end{IEEEImpStatement}


\begin{figure}[t]
\centering
\includegraphics[width=\columnwidth]{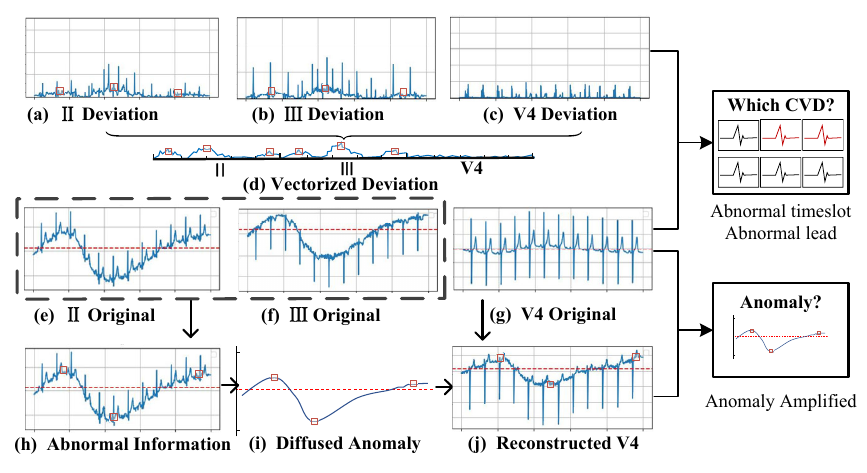}
\caption{The deviation localization and association reconstruction of multi-lead ECG data. (a)–(c) Deviation localization: per‐lead anomaly localization maps for leads II, III, and V4. (d) Vectorized deviation from continuous view to capture anomaly. (e)–(g) Original ECG waveforms for a MI patient. (h) Deviation-related contextual information derived from leads II and III.  (i) Deviation-guided cross-lead association for the target lead.  (j) Reconstructed lead V4 after association reconstruction. The red squares mark the position of obvious deviations.}
\label{fig:idea}
\end{figure}

\section{Introduction}

\IEEEPARstart{E}lectrocardiogram (ECG) signals are widely used in cardiovascular medicine, providing non-invasive, \textcolor{magenta}{real-time} insights into heart \textcolor{magenta}{function}. ECG analysis plays a key role in the assessment and management of risk of arrhythmias, myocardial infarction, heart failure, and sudden cardiac death~\cite{xiao2023deep, al2023machine, silvetti2023pivotal, muzammil2024artificial}. The increasing availability of large-scale ECG datasets from clinical systems and wearable devices has supported the application of deep learning to heartbeat classification~\cite{arslan2024ecg, avanzato2020automatic} and ECG-based heart disease classification~\cite{nguyen2021deep}. Other applications include detection of atrial fibrillation~\cite{gunduz2023atrial}, identification and risk stratification of coronary cardiovascular risk~\cite{shiraishi2023improved, hughes2023deep}. Most existing models are trained in a supervised, task-specific manner, with labels tied to particular diagnostic outcomes. These models typically learn associations specific to the training dataset, constrained by patient demographics, disease distribution, and recording conditions. As a result, learned representations often lack diversity and fail to generalize across samples, tasks, and databases.

Self-supervised representation learning uses unlabeled ECGs to learn representations of temporal patterns and inter-lead dependencies. Recent ECG representation-learning and anomaly-detection approaches include BeatGAN~\cite{zhou2019beatgan}, ASTCL~\cite{wang2023adversarial}, and ECGAD~\cite{jiang2024self}. ECG-FM~\cite{mckeen2024ecg} combines self-supervised pretraining objectives, whereas ECGFounder~\cite{Li2025ECGFounder} learns from large-scale clinical diagnostic annotations. Prior work has also explored the structure and interpretability of learned representations~\cite{mckeen2024ecg}. Foundation models developed for vision or general time-series analysis include DINOv2~\cite{oquab2023dinov2}, TimesFM~\cite{das2023decoder}, MOIRAI~\cite{woo2024unified}, and Chronos~\cite{ansari2024chronos}. In this study, we focus on lead-wise deviations relative to normal ECG morphology and their integration through deviation-guided cross-lead reconstruction.

We propose a self-supervised ECG feature extraction framework that explicitly targets the extraction of both effective and generalizable features. Unlike prior approaches that focus on building connection between ECG and cardiac disease labels, our framework learns transferable representations from unlabeled ECG signals using two structured embeddings. This enables the model to learn label-free, transferable representations applicable and better understand the meaning of vibrations across diverse diagnostic tasks.

To this end, we introduce two complementary modules that capture generalizable pathological deviations from distinct yet synergistic perspectives in ECGs. The Deviation Localization Embedding (DLE) module identifies fine-grained waveform-level deviations by learning deviations between test sample and standardized normal sample without CVDs (Fig.\ref{fig:idea}(a)-(d)). This highlights subtle anomalies that may carry diagnostic significance. The Association Reconstruction Embedding (ARE) module then uses the localized deviation cues to guide cross-lead association reconstruction through a DLE-guided gating mechanism  (Fig.\ref{fig:idea}(h)-(j)). By reweighting the contribution of peer leads during target-lead reconstruction, ARE encourages the learned representation to capture cross-lead consistency that may not be fully characterized by individual leads alone.  DLE and ARE jointly provide complementary deviation-related and cross-lead contextual information for masked reconstruction, encouraging the model to retain informative ECG representations without relying on diagnostic labels.  These features remain expressive enough to distinguish pathological signals from normal ECGs, without relying on annotated labels.

By combining DLE and ARE, our model captures beat-level anomalies and multi-lead patterns, addressing low-SNR and patient variability issues. We pretrain on unlabeled ECGs and adapt to classifiers without any fine-tuning. Our contributions are:

\begin{itemize}
  \item A self‐supervised ECG feature extractor pre‐trained on 918{,}915 recordings, learning generalizable intra-lead and cross‐lead representations.
  \item A dual-module Deviation Learning Embedding(DLE) and Association Reconstruction Embedding(ARE) framework that identifies subtle waveform deviations and integrates deviation-guided cross-lead associations for abnormally-aware ECG representation learning. 
  \item Our approach has SOTA transfer performance on four public datasets, outperforming prior self-supervised extractors and ECG classifiers without downstream adaptation.
\end{itemize}

\section{Related Work}
\subsection{Feature Representation and Extraction}

Feature extraction is central to electrocardiogram (ECG) analysis, as it converts high-dimensional physiological signals into compact and discriminative representations~\cite{kavya2015feature,alpaydin2020introduction}. Classical methods, including statistical descriptors, Fourier and wavelet transforms~\cite{gray2012fourier,gholami2012sparse}, and principal component analysis (PCA)~\cite{kurita2021principal}, rely heavily on hand-crafted features and prior signal-processing assumptions. In contrast, deep learning models, such as convolutional neural networks (CNNs)~\cite{sadouk2019cnn,jin2020prediction} and recurrent neural networks (RNNs)~\cite{wang2022ngcu,khaldi2023best}, can automatically learn task-relevant ECG embeddings from data. Their representation ability can be further improved by self-supervised pretraining and data augmentation strategies~\cite{lai2023practical}.

Recent self-supervised ECG methods have explored different pretext tasks for feature learning. Contrastive Multi-segment Coding (CMSC)~\cite{kiyasseh2021clocs} divides ECG recordings into segments and aligns their representations, while Contrastive Predictive Coding (CPC)~\cite{Oord2018RepresentationLW,Mehari2021SelfsupervisedRL} learns temporal representations by predicting future latent representations. However, recurrent models used in such frameworks may face scalability challenges when modeling long ECG sequences. Masked autoencoder-based methods further introduce temporal and lead-wise recovery objectives for ECG representation learning~\cite{Zhang2023MaeFEMA}, and ST-MEM applies random interval masking to each lead to capture spatio-temporal dependencies~\cite{Na2024GuidingMR}. Although these approaches improve local context modeling, many of them still rely strongly on inter-lead correlations or generic reconstruction objectives. Such designs may obscure lead-specific characteristics and limit the detection of subtle, syndrome-specific abnormalities. This limitation reduces their clinical utility in scenarios where cross-lead consistency and fine-grained morphological distinctions are both important.

\subsection{Self-Supervised ECG Representation Learning}

Self-supervised learning (SSL) enables representation learning without manual annotations by exploiting the inherent structure and contextual information within data~\cite{Jing2019SelfSupervisedVF}. Representative SSL frameworks include SimCLR~\cite{Chen2020ASF}, Contrastive Predictive Coding (CPC)~\cite{Oord2018RepresentationLW}, Momentum Contrast (MoCo)~\cite{He2019MomentumCF}, and Bootstrap Your Own Latent (BYOL)~\cite{Grill2020BootstrapYO}. These methods usually define pretext tasks that encourage models to learn invariant and discriminative representations, such as distinguishing augmented views of the same input or predicting future latent representations. In ECG analysis, adaptations of CPC learn temporal representations by predicting future latent representations from preceding context~\cite{Mehari2021SelfsupervisedRL}. Liu et al.~\cite{Liu2021SelfsupervisedEP} investigated self-supervised pretraining for ECG morphology recognition using ECG-specific data augmentations. Beyond contrastive learning, generative SSL methods, such as autoencoder-based frameworks, reconstruct ECG signals to preserve morphological and structural information in the learned embeddings~\cite{Gedon2021FirstST}.

Despite their effectiveness in ECG classification, existing SSL methods still face limitations in clinical generalization. Many methods are primarily optimized for within-dataset performance and remain sensitive to distribution shifts caused by differences in patient populations, acquisition devices, recording protocols, and disease distributions. Moreover, most existing frameworks focus on temporal consistency or global signal reconstruction, while lacking explicit mechanisms to model inter-lead structure and clinically meaningful abnormal patterns across 12-lead ECG recordings. As a result, they may fail to capture rare, subtle, or previously unseen abnormalities, limiting their reliability in real-world diagnostic settings.

\begin{figure*}[!t]
\centering
\includegraphics[width= 0.97\textwidth]{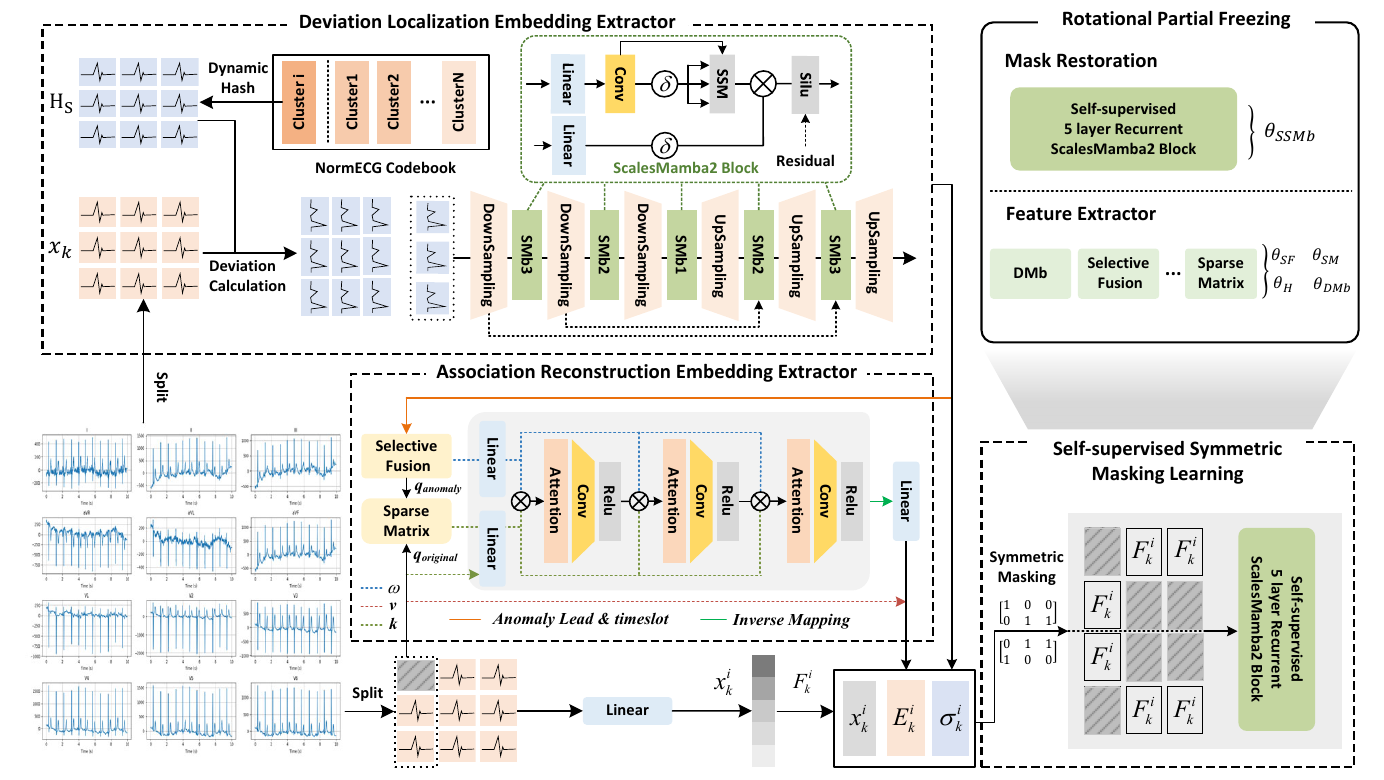}
\caption{Overview framework of ScalesMamba2-based ECG feature extraction model to learn generic dissimilarity features from unlabeled ECGs.}
\label{fig:overview}
\end{figure*}

\section{Methodology}
Our proposed framework is designed to learn robust ECG representations through a self-supervised task. The pre-training process adopts a Self-Supervised ScalesMamba2 Block (SSMb) as the backbone. The feature extraction part consists of two embedding extractors: DLE, which captures weak anomalies within ECG leads through  Deviation ScalesMamba2 (DMb), and ARE, which performs deviation-guided cross-lead association reconstruction in a latent representation space. The extracted features and the original data segment are sent to the SSMb block for mask reconstruction. The entire self-supervised framework incorporates Rotational Partial Freezing to ensure the effectiveness of the embedding extraction, see framework overview in Fig.~\ref{fig:overview}.

\subsection{ScalesMamba2 Architecture}

% To effectively model the hierarchical nature of ECG signals, where localized morphological patterns are nested within global rhythms, we introduce the ScalesMamba2 architecture. This architecture employs a U-Net topology, which excels at integrating both fine-grained details and deep contextual information. The fundamental building block of this topology is the ScalesMamba2 block, which wraps Mamba2 within a residual, multi-scale processing pattern. This block acts as a powerful feature extractor at each layer of the U-Net hierarchy. The ScalesMamba2 block is applied for its dynamic content-aware modeling ability. The rationale and design of ScalesMamba2 are provided in the supplementary material. We use a 3-layer ScalesMamba2 as DMb and a 5-layer ScalesMamba2 as SSMb. The detailed block schematics of DMb and SSMb are provided in the supplementary material.
To effectively model the hierarchical nature of ECG signals, where localized morphological patterns are nested within global rhythm dynamics, we introduce the ScalesMamba2 architecture. ECG segments contain both fine-grained P-QRS-T morphology and long-range rhythm context; therefore, the model needs to preserve local waveform details while capturing broader temporal dependencies. ScalesMamba2 adopts a U-Net-like multi-scale topology and uses Mamba2~\cite{dao2024mamba2} as its temporal modeling core.

Given an input sequence \(u_t\), Mamba2 first projects it into a hidden representation and updates an input-dependent latent state:
\begin{equation}
\begin{aligned}
h_t &= W_h u_t + b_h, \\
x_t &= A_t(h_t)x_{t-1} + B_t(h_t)h_t, \\
y_t &= C_t(h_t)x_t .
\end{aligned}
\label{eq:mamba2_state}
\end{equation}
Here, the effective state update is dynamically parameterized by the current representation \(h_t\). This content-dependent update enables the model to adapt to non-stationary ECG dynamics, including local morphological variations and rhythm changes.

The ScalesMamba2 layer wraps this temporal modeling core into a residual multi-scale operation:
\begin{equation}
\mathrm{ScalesMamba2}(U)
=
U + \mathrm{Up}\left(\mathrm{Mamba2}\left(\mathrm{Down}(U)\right)\right),
\label{eq:scalesmamba2}
\end{equation}
where \(\mathrm{Down}(\cdot)\) and \(\mathrm{Up}(\cdot)\) denote temporal resolution changes. The downsampled path allows Mamba2 to model broader temporal context on compressed representations, while the residual connection preserves fine-grained local morphological details that may be weakened during resolution changes.

Based on the ScalesMamba2 layer, we construct two U-Net-like blocks: the Deviation Mamba2 Block (DMb) and the Self-Supervised Mamba2 Block (SSMb). Both blocks stack ScalesMamba2 layers in an encoder-decoder structure with skip connections, but they are designed for different objectives. DMb is a 3-scale block used to refine the precomputed deviation signal \(\sigma\) from DLE, focusing on localized and mid-range deviation patterns. In contrast, SSMb is a deeper 5-scale block used for self-supervised masked reconstruction from raw ECG segments, where broader temporal context is required to recover missing waveform regions. This design assigns different depths according to task complexity: deviation refinement mainly focuses on deviation-related temporal patterns, whereas masked reconstruction requires both local morphology and global rhythm context.

\begin{figure*}[!t]
\centering
\includegraphics[width= 0.80\textwidth]{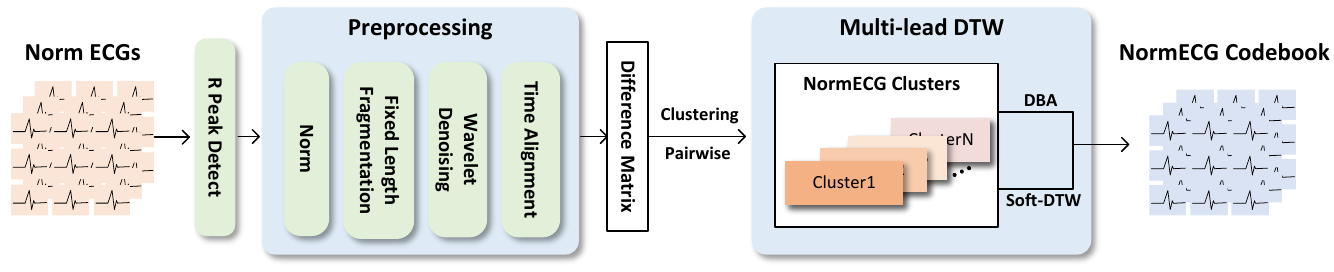}
\caption{Overall architecture of the formation of NormECG Cluster and NormECG Codebook.}
\label{fig:Codebook}
\end{figure*}

\subsection{Self-Supervised Pretraining Framework}

The model is pretrained on large-scale unlabeled ECG databases. The core objective is a deviation-aware symmetric masked reconstruction task, together with a training strategy that encourages the learning of meaningful ECG representations.

\subsubsection{Data Segmentation and Input Formulation}

Each 12-lead ECG recording is first segmented into non-overlapping windows of 500 samples, a duration typically sufficient to cover a full P-QRS-T cardiac cycle at 500 Hz. Let \(x_{k}^{(i,j)} \in \mathbb{R}^{500}\) denote the \(i\)-th segment from the \(j\)-th lead of the \(k\)-th ECG recording.

% For the reconstruction task, two embeddings are computed for each segment: \(\boldsymbol{\sigma}_{k}^{(i)}\) from DLE, which provides deviation-related cues, and \(E_{k}^{(i)}\) from ARE, which provides cross-lead association information. We then employ a symmetric masking strategy to ensure full temporal reconstruction coverage and reduce reliance on trivial local interpolation. For each lead \(j\) in segment \(i\), we generate a random binary mask \(M_{k}^{(i,j,1)}\) and its complement \(M_{k}^{(i,j,2)} = 1 - M_{k}^{(i,j,1)}\). The model input for each of the two complementary views, where \(m \in \{1,2\}\), is formed by concatenating the embeddings with the corresponding masked signal:
% \begin{equation}
%   F_{k}^{(i,j,m)} =
%   \left[
%   {\boldsymbol{\sigma}}_{k}^{(i)};
%   E_{k}^{(i)};
%   M_{k}^{(i,j,m)} \odot x_{k}^{(i,j)}
%   \right].
% \end{equation}
% This setup forces the model to leverage the learned deviation and association embeddings to reconstruct the missing parts of the signal.
For the reconstruction task, two embeddings are computed for each segment: 
\(\sigma_{k}^{(i)}\) from DLE, which provides deviation-related cues, and 
\(E_{k}^{(i)}\) from ARE, which provides cross-lead association information. 
We then employ a symmetric masking strategy to ensure full temporal reconstruction coverage 
and reduce reliance on trivial local interpolation. For each lead \(j\) in segment \(i\), 
we generate a random binary mask \(M_{k}^{(i,j,1)}\) and its complement 
\(M_{k}^{(i,j,2)} = 1 - M_{k}^{(i,j,1)}\), where \(M=1\) denotes visible positions used as input. 
Thus, each temporal position is masked and reconstructed once across the two complementary views.

The model input for each view \(m \in \{1,2\}\) is formed by concatenating the embeddings with the corresponding masked signal:
\begin{equation}
F_{k}^{(i,j,m)}
=
\left[
\sigma_{k}^{(i)};
E_{k}^{(i)};
M_{k}^{(i,j,m)} \odot x_{k}^{(i,j)}
\right].
\end{equation}
This setup encourages the model to reconstruct masked regions by leveraging both deviation-related cues and cross-lead association information.

\subsubsection{Reconstruction Loss}

The reconstruction loss is computed only on the masked-out regions. We first calculate the masked residual for each view \(m\):
\begin{equation}
  r_{k}^{(i,j,m)} =
  \left(
  \hat{x}_{k}^{(i,j,m)} - x_{k}^{(i,j)}
  \right)
  \odot
  \left(
  1 - M_{k}^{(i,j,m)}
  \right),
\end{equation}
where \(\hat{x}_{k}^{(i,j,m)}\) is the model reconstruction. The total loss is the sum of Smooth L1 losses over all dimensions, which balances robustness and smoothness:
\begin{equation}
  \mathcal{L}_{\mathrm{SS}} =
  \sum_{i,j,k,m}
  \mathcal{L}_{\delta}^{\mathrm{SmoothL1}}
  \left(
  r_{k}^{(i,j,m)}, 0
  \right).
\end{equation}

\subsubsection{Weighted Rotational Partial Freezing}

% To ensure that the model learns generalizable features rather than overfitting to the masked reconstruction objective, we employ the Weighted Rotational Partial Freezing (RPF) strategy. Additional details are provided in the supplementary material. Model parameters \(\theta\) are partitioned into blocks, separating feature extractor modules from the reconstruction head. At each step \(t\), a block \(g(t)\) is sampled for update according to a non-uniform probability distribution \(\mathbf{p}\), which assigns higher update frequencies to the feature extractor. The update rule is:
To ensure that the model learns generalizable features rather than overfitting to the masked reconstruction objective, we employ the Weighted Rotational Partial Freezing (RPF) strategy. Specifically, the model parameters \(\theta\) are partitioned into \(G\) disjoint blocks,\(\theta = [\theta_1, \dots, \theta_G]\), which are grouped into the feature extractor \(\mathcal{E}\) and self-supervision head \(\mathcal{H}\). The feature extractor contains the representation-learning modules, while the self-supervision head contains the reconstruction-specific layers responsible for masked signal recovery. At each training step \(t\), one block \(g(t)\) is selected for update according to a non-uniform probability distribution \(\mathbf{p} = [p_1, \dots, p_G]\), where \(p_g = \Pr(g(t)=g)\), while the remaining blocks are temporarily frozen. To prioritize representation learning, extractor blocks are assigned higher update probabilities than head blocks, i.e., \(p_i > p_j\) for \(\theta_i \in \mathcal{E}\) and \(\theta_j \in \mathcal{H}\). The total update probability assigned to the self-supervision head can be further bounded by \(\gamma_{\max}\), preventing the reconstruction-specific layers from dominating the optimization. The update rule is:

\begin{equation}
\label{eq:theta_update_final_v3}
  \theta^{(t+1)}
  =
  \theta^{(t)}
  -
  \alpha^{(t)}
  M_{g(t)}
  \odot
  \nabla_{\theta}\mathcal{L}_{\mathrm{SS}},
\end{equation}
where \(M_{g(t)}\) is the mask activating only block \(g(t)\). This weighted block-wise update prioritizes the learning of transferable deviation and association embeddings while regularizing the reconstruction-specific head.

\subsection{Feature Extraction Modules}

\subsubsection{Deviation Localization Embedding}

The Deviation Localization Embedding (DLE) quantifies a segment's deviation from a learned codebook of normal cardiac morphologies.

\textbf{NormECG Codebook:}
A codebook \(\mathcal{B}_{\mathrm{NormECG}} = \{\mu_c\}_{c=1}^C\) of \(C\) normal ECG prototypes is built via Bayesian clustering with Soft-DTW distance, governed by a smoothing parameter \(\lambda\). The prototypes \(\mu_c\) are updated via a standard M-step, as shown in Fig.~\ref{fig:Codebook}. The codebook serves as a morphology-level reference of normal ECG patterns. Before clustering, normal ECG segments are temporally aligned to reduce phase shifts among cardiac cycles. Multi-lead DTW and barycenter aggregation are then used to obtain representative normal prototypes, which are later used for per-lead deviation calculation.

\textbf{Deviation Calculation:}
For an input segment \(x_k^{(i)}\), we compute its similarity index \(H^S\) through an MLP layer for dynamic hashing. We then find its \(M\) nearest neighbors \(\{\mu_{c_n^*}\}_{n=1}^M\) from the codebook using cosine similarity on \(H^S\). The per-lead deviation is computed as the RMSE between the input lead and the retrieved normal prototypes:
\begin{equation}
\label{eq:sigma_final_v3}
    \sigma_{k,j}^{(i)} =
    \mathrm{RMSE}
    \left(
    x_{k,j}^{(i)} - \mu_{c_n^*,j}
    \right).
\end{equation}

\textbf{Deviation Refinement and Embedding:}
Each per-lead deviation signal \(\sigma_{k,j}^{(i)}\) is refined by a 3-scale Deviation Mamba2 Block (DMb) to capture deviation patterns across multiple temporal scales. The final embedding is the sum of these refined signals:
\begin{equation}
\label{eq:sigma_hat_final_v3}
    \boldsymbol{\sigma}_k^{(i)}
    =
    \sum_{j=1}^{L}
    \mathrm{DMb}
    \left(
    \sigma_{k,j}^{(i)}
    \right).
\end{equation}
This embedding is also passed through a scaled sigmoid function \(\tau(\cdot)\) to produce a diagonal Selective Fusion Matrix (SFM), \(\mathrm{SFM}_k^{(i)} \in \mathbb{R}^{L \times L}\), which assigns larger weights to leads with stronger deviations and provides deviation-guided lead weights for ARE

\subsubsection{Multi-Lead Association Reconstruction Embedding}

% The Multi-Lead Association Reconstruction Embedding (ARE) module reconstructs each lead from its peers, guided by the anomaly assessment from DLE. For each target lead \(j\), the following steps are performed.
The Multi-Lead Association Reconstruction Embedding (ARE) module reconstructs each target lead from its peer leads under the guidance of DLE-derived deviation cues. ARE aims to align lead representations in a shared latent association space and integrate cross-lead contextual information through association reconstruction. For each target lead \(j\), the following steps are performed.

\textbf{Context Assembly and Selection:}
% The initial context \(Q_{0,k}^{(j)}\) is formed by the other \(L-1\) leads. This context is then pruned by a composite selector matrix \(S_k^{(j)}\) to form the final context \(Q_k^{(j)} = S_k^{(j)} Q_{0,k}^{(j)}\). The selector \(S_k^{(j)}\) is constructed to: 1) enforce a sparse, structured information flow using a learnable sparse matrix \(M^{(j)}\), and 2) always include leads that DLE has flagged as anomalous via the \(\mathrm{SFM}_k^{(i)}\) matrix:
The initial context \(Q_{0,k}^{(j)}\) is formed by the other \(L-1\) leads. This context is then pruned by a composite selector matrix \(S_k^{(j)}\) to form the final context \(Q_k^{(j)} = S_k^{(j)} Q_{0,k}^{(j)}\). The selector \(S_k^{(j)}\) is constructed to: 1) capture sparse and structured lead-to-lead associations using a learnable sparse matrix \(M^{(j)}\), and 2) incorporate peer leads with stronger deviation cues emphasized by the DLE-derived \(\mathrm{SFM}_k^{(i)}\) matrix:
\begin{equation}
\label{eq:D_S_F_final_v3}
    S_k^{(j)}
    =
    M^{(j)}
    \lor
    \mathrm{diag}
    \left(
    \mathbf{1}_{\{
    \mathrm{diag}
    (
    \mathrm{SFM}_k^{(\setminus j)}
    )
    > 0
    \}}
    \right).
\end{equation}
Here, \(\mathrm{SFM}_k^{(\setminus j)}\) denotes the segment-level aggregated \(\mathrm{SFM}_k\) excluding lead \(j\).

\textbf{Deviation-Guided Attention:}
% The context \(Q_k^{(j)}\) is processed by attention layers. To bridge the DLE and ARE modules, we introduce an anomaly-gating mechanism that uses the deviation signal to modulate the attention scores. The key vectors are amplified based on the per-lead deviations of the context leads:
The context \(Q_k^{(j)}\) is processed by attention layers. To bridge the DLE and ARE modules, we introduce a deviation-guided gating mechanism that uses the DLE-derived SFM to modulate the attention computation. The key vectors are reweighted according to the per-lead deviation cues of the context leads::
\begin{equation}
\label{eq:QKV_final_v3}
    K_t^{(j)}
    =
    \left(
    I + \beta \mathrm{SFM}_k^{(\setminus j)}
    \right)
    W_t^K
    Q_k^{(j)}.
\end{equation}
% This forces the model to focus on context from other leads that are also anomalous, enabling effective cross-lead propagation of anomaly information. The attention output is then computed as
This guides the attention module to align the target-lead representation with peer-lead representations carrying stronger deviation cues, enabling the reconstruction process to model cross-lead consistency in the latent association space. The attention output is then computed as:
\begin{equation}
\label{eq:attention_final_v3}
    Z_t^{(j)}
    =
    \mathrm{Attn}
    \left(
    Q_k^{(j)},
    K_k^{(j)},
    V_k^{(j)}
    \right).
\end{equation}

\textbf{Reconstruction and Fusion:}
% The attention output \(Z_t^{(j)}\) is used to generate a reconstruction \(\hat{x}_{k,j}\) of the target lead \(j\). The final ARE embedding \(E_k\) is the aggregation of all such per-lead reconstructions, producing a robust, denoised, and contextually aware representation of the entire segment:
The attention output \(Z_t^{(j)}\) is used to generate a reconstruction \(\hat{x}_{k,j}\) of the target lead \(j\). The final ARE embedding \(E_k\) is the aggregation of all such per-lead reconstructions, producing a context-aware embedding that integrates aligned cross-lead representations while reducing isolated lead-specific fluctuations:
\begin{equation}
\label{eq:reconstruct_final_v3}
    E_k^{(i)}
    =
    \sum_{j=1}^{L}
    \hat{x}_{k,j}.
\end{equation}

\begin{table*}[!t]
\centering
\caption{Classification performance of different downstream classifiers on PTB-XL and ML12L using 10-fold cross-validation, and on CPSC2020 and WGEX using external validation. Untokenized denotes models trained directly on raw ECG signals without a feature extraction framework. The best result in each column is shown in \textbf{bold}, the second-best is \underline{underlined}, and the worst is \textit{italicized}.}
\label{tab:linear_eval}

\scriptsize
\setlength{\tabcolsep}{3.2pt}
\renewcommand{\arraystretch}{0.98}

\begin{tabular*}{\textwidth}{@{\extracolsep{\fill}}llcccccc}
\toprule
\textbf{Classifier} & \textbf{Extractor}
& \multicolumn{3}{c}{\textbf{PTB-XL}}
& \multicolumn{3}{c}{\textbf{ML12L}} \\
\cmidrule(lr){3-5}\cmidrule(lr){6-8}
& & \textbf{B-Acc} & \textbf{F1} & \textbf{AUROC}
  & \textbf{B-Acc} & \textbf{F1} & \textbf{AUROC} \\
\midrule

\multirow{6}{*}{CNN}
  & Untokenized & 0.399$\pm$0.024 & 0.338$\pm$0.042 & 0.694$\pm$0.011
                & 0.701$\pm$0.162 & \underline{0.425$\pm$0.056} & 0.661$\pm$0.069 \\
  & Dinov2      & 0.452$\pm$0.014 & \underline{0.383$\pm$0.040} & \underline{0.757$\pm$0.006}
                & \underline{0.740$\pm$0.118} & 0.368$\pm$0.055 & \underline{0.725$\pm$0.137} \\
  & TimesFM     & \underline{0.474$\pm$0.026} & 0.225$\pm$0.053 & 0.730$\pm$0.007
                & 0.684$\pm$0.018 & 0.352$\pm$0.048 & 0.669$\pm$0.051 \\
  & Chronos     & \textit{0.382$\pm$0.055} & 0.162$\pm$0.071 & 0.597$\pm$0.008
                & 0.681$\pm$0.082 & 0.306$\pm$0.045 & 0.723$\pm$0.065 \\
  & MOIRAI      & 0.387$\pm$0.115 & \textit{0.075$\pm$0.059} & \textit{0.562$\pm$0.084}
                & \textit{0.487$\pm$0.243} & \textit{0.175$\pm$0.060} & \textit{0.562$\pm$0.237} \\
  & \textbf{Ours}
                & \textbf{0.493$\pm$0.040} & \textbf{0.511$\pm$0.017} & \textbf{0.795$\pm$0.007}
                & \textbf{0.771$\pm$0.015} & \textbf{0.440$\pm$0.045} & \textbf{0.719$\pm$0.075} \\

\midrule

\multirow{6}{*}{Mamba}
  & Untokenized & \underline{0.498$\pm$0.033} & \underline{0.492$\pm$0.024} & \underline{0.797$\pm$0.022}
                & 0.700$\pm$0.168 & \underline{0.424$\pm$0.058} & 0.679$\pm$0.069 \\
  & Dinov2      & 0.480$\pm$0.014 & 0.302$\pm$0.047 & 0.767$\pm$0.015
                & \underline{0.716$\pm$0.157} & 0.377$\pm$0.055 & 0.658$\pm$0.149 \\
  & TimesFM     & 0.477$\pm$0.015 & 0.259$\pm$0.039 & 0.741$\pm$0.007
                & 0.570$\pm$0.041 & \textit{0.342$\pm$0.039} & 0.680$\pm$0.150 \\
  & Chronos     & 0.455$\pm$0.013 & \textit{0.025$\pm$0.008} & 0.622$\pm$0.014
                & 0.772$\pm$0.010 & 0.343$\pm$0.008 & \underline{0.734$\pm$0.033} \\
  & MOIRAI      & \textit{0.198$\pm$0.030} & 0.255$\pm$0.017 & \textit{0.520$\pm$0.005}
                & \textit{0.298$\pm$0.087} & 0.355$\pm$0.014 & \textit{0.331$\pm$0.008} \\
  & \textbf{Ours}
                & \textbf{0.552$\pm$0.017} & \textbf{0.556$\pm$0.013} & \textbf{0.814$\pm$0.009}
                & \textbf{0.785$\pm$0.012} & \textbf{0.428$\pm$0.028} & \textbf{0.758$\pm$0.145} \\

\midrule

\multirow{6}{*}{Transformer}
  & Untokenized & 0.448$\pm$0.027 & 0.314$\pm$0.034 & 0.686$\pm$0.008
                & 0.548$\pm$0.035 & 0.414$\pm$0.039 & 0.566$\pm$0.068 \\
  & Dinov2      & 0.471$\pm$0.014 & 0.279$\pm$0.028 & 0.736$\pm$0.007
                & 0.571$\pm$0.045 & 0.379$\pm$0.048 & 0.638$\pm$0.110 \\
  & TimesFM     & \underline{0.479$\pm$0.019} & \underline{0.338$\pm$0.069} & \underline{0.767$\pm$0.023}
                & \underline{0.579$\pm$0.025} & \underline{0.438$\pm$0.056} & \underline{0.673$\pm$0.040} \\
  & Chronos     & 0.362$\pm$0.025 & \textit{0.218$\pm$0.028} & \textit{0.591$\pm$0.010}
                & 0.462$\pm$0.011 & \textit{0.318$\pm$0.042} & \textit{0.403$\pm$0.054} \\
  & MOIRAI      & \textit{0.331$\pm$0.021} & 0.285$\pm$0.021 & 0.611$\pm$0.011
                & \textit{0.431$\pm$0.023} & 0.385$\pm$0.021 & \textit{0.416$\pm$0.148} \\
  & \textbf{Ours}
                & \textbf{0.520$\pm$0.023} & \textbf{0.507$\pm$0.033} & \textbf{0.797$\pm$0.008}
                & \textbf{0.620$\pm$0.040} & \textbf{0.607$\pm$0.044} & \textbf{0.715$\pm$0.061} \\

\midrule[0.8pt]

& & \multicolumn{3}{c}{\textbf{CPSC2020}}
  & \multicolumn{3}{c}{\textbf{WGEX}} \\
\cmidrule(lr){3-5}\cmidrule(lr){6-8}
& & \textbf{B-Acc} & \textbf{F1} & \textbf{AUROC}
  & \textbf{B-Acc} & \textbf{F1} & \textbf{AUROC} \\
\midrule

\multirow{6}{*}{CNN}
  & Untokenized & 0.356$\pm$0.046 & 0.385$\pm$0.041 & 0.696$\pm$0.020
                & 0.485$\pm$0.104 & \underline{0.431$\pm$0.041} & 0.569$\pm$0.031 \\
  & Dinov2      & \underline{0.483$\pm$0.089} & 0.372$\pm$0.057 & \underline{0.760$\pm$0.031}
                & \underline{0.512$\pm$0.091} & 0.404$\pm$0.041 & 0.472$\pm$0.048 \\
  & TimesFM     & 0.464$\pm$0.013 & \underline{0.379$\pm$0.073} & 0.743$\pm$0.024
                & 0.509$\pm$0.075 & 0.364$\pm$0.046 & 0.447$\pm$0.047 \\
  & Chronos     & 0.450$\pm$0.020 & \textit{0.046$\pm$0.060} & \textit{0.593$\pm$0.043}
                & 0.444$\pm$0.033 & 0.320$\pm$0.034 & \underline{0.816$\pm$0.049} \\
  & MOIRAI      & 0.322$\pm$0.060 & 0.162$\pm$0.024 & 0.618$\pm$0.020
                & 0.322$\pm$0.060 & \textit{0.262$\pm$0.024} & 0.439$\pm$0.045 \\
  & \textbf{Ours}
                & \textbf{0.534$\pm$0.002} & \textbf{0.505$\pm$0.017} & \textbf{0.800$\pm$0.004}
                & \textbf{0.521$\pm$0.009} & \textbf{0.478$\pm$0.010} & \textbf{0.798$\pm$0.070} \\

\midrule

\multirow{6}{*}{Mamba}
  & Untokenized & 0.478$\pm$0.038 & \underline{0.460$\pm$0.057} & 0.770$\pm$0.037
                & 0.494$\pm$0.006 & \underline{0.423$\pm$0.057} & 0.654$\pm$0.086 \\
  & Dinov2      & \underline{0.488$\pm$0.097} & 0.333$\pm$0.026 & \underline{0.776$\pm$0.045}
                & 0.529$\pm$0.033 & 0.418$\pm$0.027 & 0.493$\pm$0.073 \\
  & TimesFM     & 0.450$\pm$0.021 & 0.167$\pm$0.057 & 0.700$\pm$0.016
                & 0.516$\pm$0.046 & 0.378$\pm$0.113 & 0.597$\pm$0.130 \\
  & Chronos     & 0.450$\pm$0.008 & \textit{0.060$\pm$0.007} & 0.626$\pm$0.008
                & \underline{0.570$\pm$0.005} & 0.342$\pm$0.004 & \underline{0.722$\pm$0.005} \\
  & MOIRAI      & \textit{0.263$\pm$0.027} & 0.211$\pm$0.021 & \textit{0.519$\pm$0.007}
                & \textit{0.363$\pm$0.027} & 0.311$\pm$0.021 & \textit{0.327$\pm$0.005} \\
  & \textbf{Ours}
                & \textbf{0.547$\pm$0.000} & \textbf{0.557$\pm$0.011} & \textbf{0.780$\pm$0.007}
                & \textbf{0.584$\pm$0.010} & \textbf{0.426$\pm$0.003} & \textbf{0.775$\pm$0.140} \\

\midrule

\multirow{6}{*}{Transformer}
  & Untokenized & 0.478$\pm$0.014 & 0.299$\pm$0.102 & 0.700$\pm$0.006
                & 0.378$\pm$0.014 & 0.399$\pm$0.102 & 0.606$\pm$0.032 \\
  & Dinov2      & \underline{0.487$\pm$0.045} & 0.250$\pm$0.037 & 0.752$\pm$0.009
                & \underline{0.387$\pm$0.045} & 0.350$\pm$0.037 & 0.622$\pm$0.025 \\
  & TimesFM     & 0.464$\pm$0.011 & \underline{0.343$\pm$0.014} & \underline{0.676$\pm$0.027}
                & 0.364$\pm$0.011 & \underline{0.443$\pm$0.014} & \underline{0.652$\pm$0.035} \\
  & Chronos     & 0.388$\pm$0.020 & \textit{0.176$\pm$0.012} & \textit{0.591$\pm$0.009}
                & 0.288$\pm$0.020 & \textit{0.276$\pm$0.012} & 0.452$\pm$0.005 \\
  & MOIRAI      & \textit{0.314$\pm$0.044} & 0.302$\pm$0.021 & 0.603$\pm$0.013
                & \textit{0.314$\pm$0.044} & 0.402$\pm$0.021 & \textit{0.402$\pm$0.045} \\
  & \textbf{Ours}
                & \textbf{0.533$\pm$0.035} & \textbf{0.548$\pm$0.002} & \textbf{0.791$\pm$0.007}
                & \textbf{0.433$\pm$0.035} & \textbf{0.538$\pm$0.002} & \textbf{0.809$\pm$0.019} \\

\bottomrule
\end{tabular*}
\end{table*}

% \begin{table}[t]
% \centering
% \caption{Performance comparison of SOTA self-supervised ECG classifiers on the unseen test sets \textbf{CPSC2020} and \textbf{WGEX} using 10-Fold Validation. }
% \label{tab:sota_comparison}
% \setlength{\tabcolsep}{2pt}
% \renewcommand{\arraystretch}{0.75} 
% \scriptsize
% \begin{tabular}{lcccc cccc}
% \toprule
% & \multicolumn{4}{c}{\textbf{CPSC2020}} & \multicolumn{4}{c}{\textbf{WGEX}} \\
% \cmidrule(lr){2-5} \cmidrule(lr){6-9}
% Model & B-Acc & F1 & AUC & Sensitivity & B-Acc & F1 & AUC & Sensitivity \\
% \midrule
% ASTCL            & 0.496 & 0.522 & 0.744 & 0.700 & 0.530 & 0.462 & 0.724 & 0.699 \\
% BeatGAN      & 0.480 & 0.546 & 0.762 & 0.690 & 0.501 & 0.476 & 0.765 & 0.711 \\
% MERL     & 0.243 & 0.462 & 0.724 & 0.686 & 0.208 & 0.260 & 0.673 & 0.522 \\
% ECGAD        & 0.502 & 0.436 & 0.740 & 0.723 & 0.515 & 0.338 & 0.730 & 0.667 \\
% \textbf{Ours}   & \textbf{0.547} & \textbf{0.557} & \textbf{0.800} & \textbf{0.740} & \textbf{0.584} & \textbf{0.538} & \textbf{0.809} & \textbf{0.712} \\
% \bottomrule
% \end{tabular}
% \end{table}

\begin{table}[t]
\centering
\caption{Performance comparison of SOTA self-supervised ECG classifiers on the unseen test sets \textbf{CPSC2020} and \textbf{WGEX} using 10-fold validation.}
\label{tab:sota_comparison}

\small
\setlength{\tabcolsep}{0pt}
\renewcommand{\arraystretch}{1.08}

\begin{tabular*}{\columnwidth}{@{\extracolsep{\fill}}lcccccccc@{}}
\toprule
\textbf{Model}
& \multicolumn{4}{c}{\textbf{CPSC2020}}
& \multicolumn{4}{c}{\textbf{WGEX}} \\
\cmidrule(lr){2-5}
\cmidrule(lr){6-9}
& \textbf{B-Acc} & \textbf{F1} & \textbf{AUC} & \textbf{Sens.}
& \textbf{B-Acc} & \textbf{F1} & \textbf{AUC} & \textbf{Sens.} \\
\midrule
ASTCL
& 0.496 & 0.522 & 0.744 & 0.700
& 0.530 & 0.462 & 0.724 & 0.699 \\

BeatGAN
& 0.480 & 0.546 & 0.762 & 0.690
& 0.501 & 0.476 & 0.765 & 0.711 \\

MERL
& 0.243 & 0.462 & 0.724 & 0.686
& 0.208 & 0.260 & 0.673 & 0.522 \\

ECGAD
& 0.502 & 0.436 & 0.740 & 0.723
& 0.515 & 0.338 & 0.730 & 0.667 \\

\textbf{Ours}
& \textbf{0.547} & \textbf{0.557} & \textbf{0.800} & \textbf{0.740}
& \textbf{0.584} & \textbf{0.538} & \textbf{0.809} & \textbf{0.712} \\
\bottomrule
\end{tabular*}

\vspace{0.5mm}
\footnotesize
\textit{Sens. denotes sensitivity.}
\end{table}

\section{Experiments and Results}

\subsection{Datasets}

\textbf{Pretraining data.}
We pretrained our feature extractor on 918{,}915 ECG recordings, comprising 800{,}137 from the MIMIC-IV-ECG~\cite{gow2023mimicivecg} and 118{,}778 from a clinical partner. The data were randomly split into 80\% for training and 20\% for validation. 

\textbf{Evaluation datasets.}
We used four publicly available 12-lead ECG datasets for validation experiments. \textbf{PTB-XL}~\cite{wagner2020ptb} contains 21{,}837 10-second recordings at 500~Hz annotated into five diagnostic classes from Europe. \textbf{ML12L}~\cite{liu2022large} includes 25{,}770 recordings with durations of 10--60~s at 500~Hz, grouped into eight multi-label classes from China. \textbf{CPSC2020}~\cite{alday2020classification} provides 10{,}330 recordings with durations of 6--60~s at 500~Hz, labeled across nine classes from China. \textbf{WGEX}~\cite{zheng2022large} comprises 12{,}600 10-second recordings at 500~Hz spanning seven classes from China. Among them, PTB-XL and ML12L are used for frozen-feature evaluation, while CPSC2020 and WGEX are utilized for external generalizability comparison.

\begin{figure*}[t]
\centering
\includegraphics[width=0.97\textwidth]{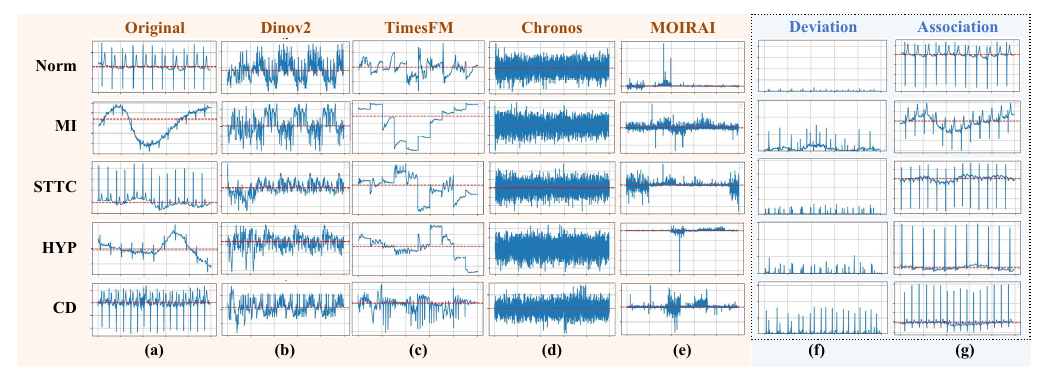}
\caption{Visualization of extracted features using different models. Different columns show features extracted by different models. Original denotes the raw ECG signals, while Deviation and Association denote features from the DLE and ARE extractors, respectively.}
\label{fig:featurevisualization}
\end{figure*}

\subsection{Models for Comparison}

We compared our method against state-of-the-art self-supervised feature extractors for frozen-feature evaluation and state-of-the-art zero-shot ECG classification models for generalization validation.

The compared models in frozen-feature evaluation include \textbf{DINOv2}~\cite{oquab2023dinov2}, \textbf{TimesFM}~\cite{das2023decoder}, \textbf{Chronos}~\cite{ansari2024chronos}, and \textbf{MOIRAI}~\cite{woo2024unified}. The models used in the generalizability comparison include \textbf{ASTCL}~\cite{wang2023adversarial}, \textbf{BeatGAN}~\cite{zhou2019beatgan}, \textbf{MERL}~\cite{liu2024zero}, and \textbf{ECGAD}~\cite{jiang2024self}. 
\subsection{Implementation Details}

Our experiments used 24 V100 GPUs to train self-supervised models, run benchmark comparisons, and perform ablation studies. Pretrained parameters for DINOv2, TimesFM, Chronos, and MOIRAI were configured for large-scale deployment in frozen-feature evaluation. The zero-shot classifiers, e.g., ASTCL, BeatGAN, MERL, and ECGAD, were loaded with large-scale parameter tables and trained using the same parameter settings. All configurations and hyperparameter setups are consistent with our model.

For self-supervised masking training, we partitioned the pretraining datasets into 100 subsets. For each subset, we tracked the best-performing parameters based on validation loss and propagated them to the next subset, enabling a continuous training process. For downstream evaluation, all datasets followed a 10-fold cross-validation protocol. PTB-XL and ML12L were used as benchmark datasets, while CPSC2020 and WGEX served as external test sets. The downstream learning rate was initialized at \(1 \times 10^{-4}\) and halved every 5 epochs over 20 epochs. To isolate the effect of feature representations, we conducted a frozen-feature evaluation by feeding different extracted features into the same downstream classifiers. The downstream classifiers contain three different encoders with different structures, including a four-layer convolutional network, a Transformer with four encoder layers and attention heads, and a Mamba-based sequence model, followed by the same linear decoder. 

\begin{figure*}[t]
\centering
\includegraphics[width=\textwidth]{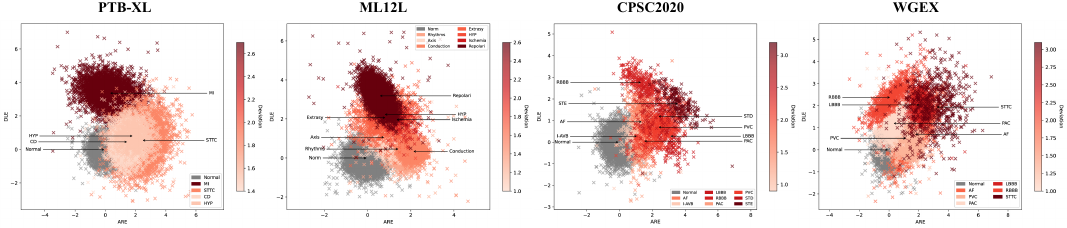}
\caption{Two-dimensional projection of extracted features of DLE and ARE on PTB-XL, ML12L, CPSC2020, and WGEX datasets. The X-axis shows the ARE mapping, and the Y-axis shows the DLE mapping. Gray points represent healthy patients; points in other colors correspond to different cardiac diseases.}
\label{fig:featureMap}
\end{figure*}

\subsection{Comparison with State-of-the-Art Models}

We first conducted a frozen-feature evaluation by training lightweight classifiers, including CNN, Mamba, and Transformer, on feature representations extracted from different pretrained models. As shown in Table~\ref{tab:linear_eval}, our method consistently achieved the best balanced accuracy, F1 score, and AUROC on both PTB-XL and ML12L.

With downstream training, our features outperformed all baselines on PTB-XL and ML12L. On PTB-XL, under the CNN classifier, our method achieved superior performance compared with TimesFM. Under the Mamba classifier, our method also outperformed the strongest baseline. On ML12L, our method achieved better performance under the Transformer classifier, exceeding TimesFM by 4.1\% in F1 score and 4.2\% in AUROC. When the pretrained feature extractor was frozen and evaluated on external datasets, including CPSC2020 and WGEX, our model also showed the best overall performance. Under the Transformer classifier, our model exceeded TimesFM by over 20\% in F1 score. These results demonstrate the robustness of our learned features across different classifiers and their resilience to label imbalance in multi-label tasks.

Our model showed better performance than other methods on unseen datasets. We evaluated its generalizability to out-of-distribution data using CPSC2020 and WGEX. As shown in Table~\ref{tab:sota_comparison}, our approach consistently outperformed all baselines. On CPSC2020, it exceeded ECGAD by 4.5--6.4\% across B-Acc, F1 score, AUROC, and sensitivity. On WGEX, which uses a different acquisition protocol and sampling rate, it surpassed BeatGAN by 4.4--6.3\% across the same metrics.

Overall, our approach delivers transferable ECG features across models, label settings, and datasets. Its consistent performance suggests that the proposed pretraining strategy captures both local waveform deviations and lead-level associations more effectively than prior methods.

\subsection{Generated Embedding Visualization Analysis}

Fig.~\ref{fig:featurevisualization} compares the structure of raw ECG waveforms and the embeddings learned by different methods. Fig.~\ref{fig:featurevisualization}(a) shows the raw signals, which display different waveforms among normal (Norm), myocardial infarction (MI), ST-T changes (STTC), hypertrophy (HYP), and conduction disorder (CD) classes, without any anomaly indicators. DINOv2 in Fig.~\ref{fig:featurevisualization}(b) and TimesFM in Fig.~\ref{fig:featurevisualization}(c) partially separate some classes but exhibit considerable inter-class mixing. Chronos in Fig.~\ref{fig:featurevisualization}(d) and MOIRAI in Fig.~\ref{fig:featurevisualization}(e) produce fragmented and inconsistent groupings, reflecting weak discriminative capability.

In contrast, our deviation localization embedding in Fig.~\ref{fig:featurevisualization}(f) forms compact and class-consistent clusters that clearly distinguish pathological patterns from normal ones. The association reconstruction embedding in Fig.~\ref{fig:featurevisualization}. (g) further improves class separation by integrating deviation-guided cross-lead contextual information, yielding more compact and distinct clusters.  Finally, the NormECG codebook projection outlines the reference manifold of normal segments used to compute per-lead deviations, as detailed in the supplementary material. These qualitative patterns are consistent with the quantitative improvements reported in Table~\ref{tab:linear_eval} for frozen-feature evaluation on all datasets and Table~\ref{tab:sota_comparison} for generalizability comparison on external datasets, confirming that our learned representations are both more robust and more discriminative.

\begin{table*}[!t]
\centering
\caption{Classification performance of different downstream classifiers using Baseline and DLE\&ARE as ECG representations on external clinical datasets. The best result in each column is shown in \textbf{bold}.}
\label{tab:observational_study_result}
\scriptsize
\setlength{\tabcolsep}{6pt}
\renewcommand{\arraystretch}{1.05}
\begin{tabular}{llccccc}
\toprule
\textbf{Classifier} & \textbf{Representation} 
& \textbf{Accuracy} & \textbf{F1 Score} & \textbf{Precision} & \textbf{Recall} & \textbf{AUC} \\
\midrule

\multirow{2}{*}{CNN}
& Baseline
& \(0.221\pm0.011\) & \(0.356\pm0.012\) & \(0.449\pm0.014\) & \(0.303\pm0.018\) & \(0.573\pm0.015\) \\
& DLE\&ARE
& \(\textbf{0.248}\pm\textbf{0.012}\) & \(\textbf{0.395}\pm\textbf{0.016}\) & \(\textbf{0.466}\pm\textbf{0.009}\) & \(\textbf{0.348}\pm\textbf{0.022}\) & \(\textbf{0.602}\pm\textbf{0.011}\) \\
\midrule

\multirow{2}{*}{Transformer}
& Baseline
& \(0.242\pm0.024\) & \(0.352\pm0.024\) & \(\textbf{0.532}\pm\textbf{0.009}\) & \(0.269\pm0.026\) & \(0.574\pm0.011\) \\
& DLE\&ARE
& \(\textbf{0.276}\pm\textbf{0.027}\) & \(\textbf{0.401}\pm\textbf{0.022}\) & \(0.512\pm0.011\) & \(\textbf{0.341}\pm\textbf{0.030}\) & \(\textbf{0.588}\pm\textbf{0.007}\) \\
\midrule

\multirow{2}{*}{Mamba}
& Baseline
& \(0.241\pm0.017\) & \(0.386\pm0.022\) & \(0.399\pm0.021\) & \(0.377\pm0.025\) & \(0.499\pm0.003\) \\
& DLE\&ARE
& \(\textbf{0.348}\pm\textbf{0.029}\) & \(\textbf{0.429}\pm\textbf{0.028}\) & \(\textbf{0.524}\pm\textbf{0.008}\) & \(\textbf{0.377}\pm\textbf{0.031}\) & \(\textbf{0.543}\pm\textbf{0.009}\) \\
\bottomrule
\end{tabular}
\end{table*}

\begin{table}[!t]
\centering
\caption{Ablation study on CPSC2020 and WGEX external validation. SSMb, SymMask, RPF, DLE, DMb, and ARE denote the Self-supervised 5-layer ScalesMamba2 pretraining backbone, Symmetric Masking Strategy, Weighted Rotational Partial Freezing, Deviation Localization Embedding, Deviation Mamba2 Block, and Association Reconstruction Embedding, respectively. A checkmark indicates inclusion of the component; a blank indicates omission.}
\label{tab:ablation_study}
\scriptsize
\setlength{\tabcolsep}{2pt}
\renewcommand{\arraystretch}{1.05}
\resizebox{\columnwidth}{!}{
\begin{tabular}{cccccccccccc}
\toprule
\textbf{SSMb} & \textbf{SymMask} & \textbf{RPF} 
& \textbf{DLE} & \textbf{DMb} & \textbf{ARE}
& \multicolumn{3}{c}{\textbf{CPSC2020}} 
& \multicolumn{3}{c}{\textbf{WGEX}} \\
\cmidrule(lr){7-9} \cmidrule(lr){10-12}
& & & & & 
& B-Acc & F1 & AUC   
& B-Acc & F1 & AUC \\
\midrule
& & & $\checkmark$ & $\checkmark$ & $\checkmark$
& 0.289 & 0.295 & 0.552 
& 0.270 & 0.285 & 0.557 \\

$\checkmark$ & & & $\checkmark$ & $\checkmark$ & $\checkmark$
& 0.397 & 0.403 & 0.659 
& 0.337 & 0.390 & 0.663 \\

$\checkmark$ & $\checkmark$ & & $\checkmark$ & $\checkmark$ & $\checkmark$
& 0.466 & 0.470 & 0.726
& 0.464 & 0.456 & 0.730 \\

$\checkmark$ & $\checkmark$ & $\checkmark$ & & & $\checkmark$
& 0.454 & 0.458 & 0.712 
& 0.432 & 0.452 & 0.716 \\

$\checkmark$ & $\checkmark$ & $\checkmark$ & $\checkmark$ & $\checkmark$ &
& 0.473 & 0.475 & 0.729 
& 0.449 & 0.458 & 0.732 \\

$\checkmark$ & $\checkmark$ & $\checkmark$ & $\checkmark$ & & $\checkmark$
& 0.500 & 0.501 & 0.755
& 0.476 & 0.462 & 0.748 \\

$\checkmark$ & $\checkmark$ & $\checkmark$ & $\checkmark$ & $\checkmark$ & $\checkmark$
& \textbf{0.538} & \textbf{0.536} & \textbf{0.763}
& \textbf{0.513} & \textbf{0.471} & \textbf{0.787} \\
\bottomrule
\end{tabular}
}
\end{table}

\subsection{Embedding Space Interpretability Analysis}

% Furthermore, we analyze the generalizability of DLE and ARE embeddings for ECG anomaly grouping. Fig.~\ref{fig:featureMap} maps the two embeddings from different datasets onto different feature spaces, examining their distributional category patterns. The full figure is provided in the supplementary material.
Furthermore, we analyze the interpretability and cross-dataset consistency of the learned DLE and ARE embeddings. DLE reflects lead-wise deviations from normal ECG morphology, whereas ARE captures deviation-guided cross-lead associations in the latent representation space. Fig.~\ref{fig:featureMap} visualizes their joint distributions across different datasets to examine whether the learned embeddings form consistent and separable structures for normal and abnormal ECG categories. 

\textbf{DLE Embedding.}
DLE encodes the magnitude of deviation of each lead from the NormECG codebook. In the projected space, samples with pronounced cardiovascular abnormalities, such as ventricular ectopy and atrial fibrillation, form tight and well-defined clusters. In contrast, cases with subtle deviations, such as mild ST-segment elevation, appear more dispersed with overlapping boundaries. This pattern indicates that DLE alone provides limited discriminative power for weaker anomalies.

\textbf{ARE Embedding.}
% ARE captures cross-lead relationships by diffusing localized deviations through attention-based reconstruction. Compared with DLE, ARE yields higher intra-class compactness and better inter-class separability across all datasets. In ML12L and CPSC2020 in particular, class clusters are clearly delineated, highlighting the discriminative capability of ARE.
ARE captures deviation-guided cross-lead associations through attention-based reconstruction in the latent representation space. Compared with DLE, ARE produces more compact intra-class distributions and clearer inter-class separation across all datasets. In particular, class clusters in ML12L and CPSC2020 are more clearly delineated, indicating that ARE improves the discriminative structure of the learned representations.

\textbf{Cross-Dataset Consistency and Classification Insight.}
Across all four datasets, the embeddings separate normal and abnormal cases and distinguish among major pathological classes. This consistency demonstrates the model's ability to generalize deviation detection and localization across diverse clinical sources. Moreover, cluster tightness correlates with downstream classification performance: datasets with more cohesive and well-separated embeddings tend to yield higher detection accuracy. These visualizations thus validate the discriminative quality of our features and provide practical guidance for architecture design, including attention mechanisms and hyperparameter tuning.

\subsection{Analysis of DLE-Guided Lead Associations}
\begin{figure*}[t]
\centering
\includegraphics[width=\textwidth]{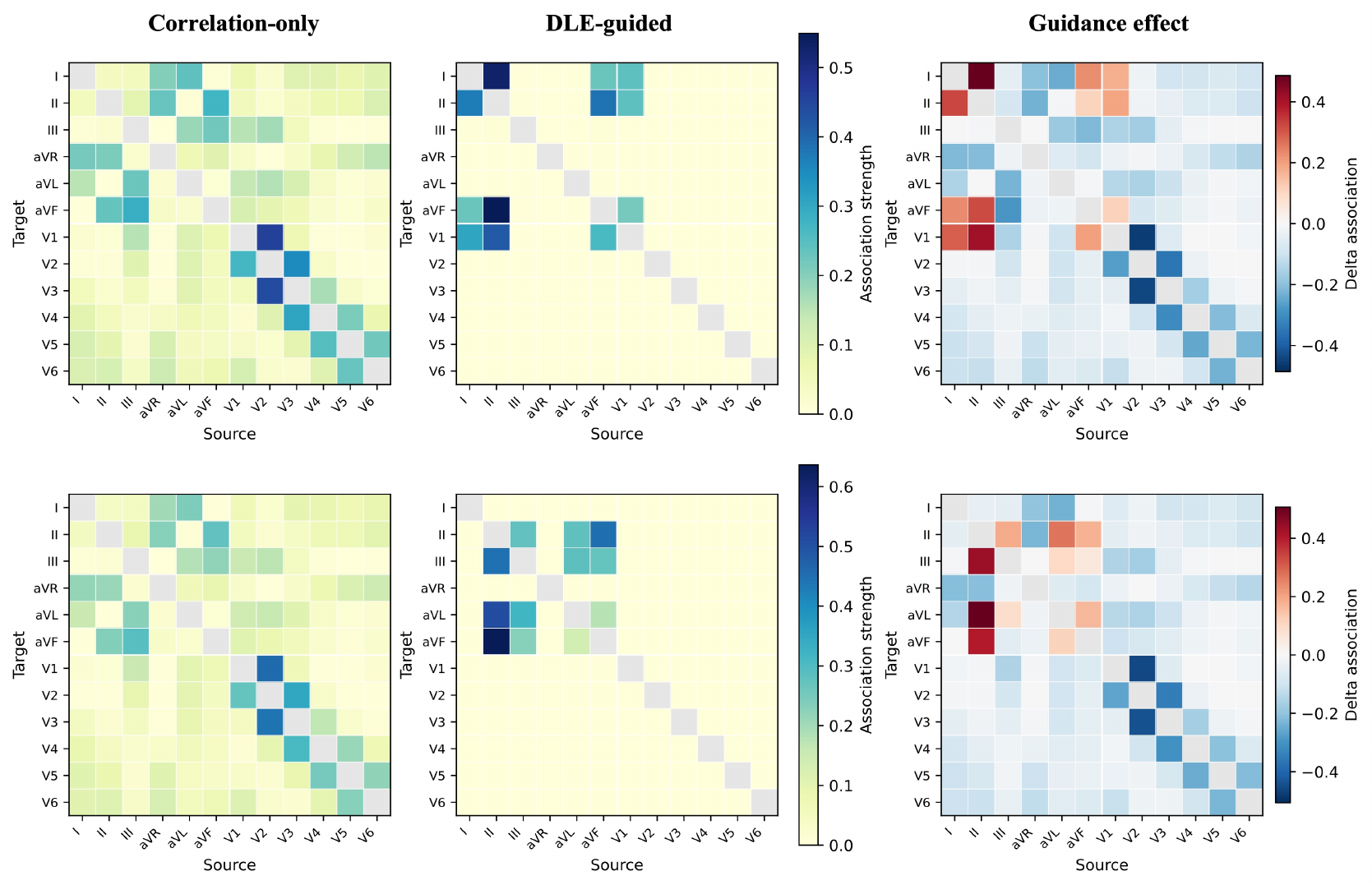}
\caption{DLE-guided lead association analysis in ARE. 
Rows and columns denote target and source leads, respectively. 
The three columns show the correlation-only reference, DLE-guided ARE association, and guidance-induced change. 
Positive and negative values indicate enhanced and suppressed source-to-target associations after DLE/SFM guidance; gray diagonal entries are excluded from cross-lead interpretation.}
\label{fig:are_guidance}
\end{figure*}

To analyze the role of DLE in ARE, we visualize the source-to-target lead associations used during cross-lead reconstruction. Fig.~\ref{fig:are_guidance} compares a correlation-only reference map with the DLE-guided association map produced by ARE. Rows correspond to target leads to be reconstructed, and columns correspond to source leads used as reconstruction context. Each entry represents the association strength from a source lead to a target lead. Diagonal entries are masked because self-lead dependencies are not considered cross-lead associations.

The correlation-only reference reflects the pairwise lead coupling estimated directly from the input ECG signals. Its association pattern is relatively diffuse, indicating that multiple limb and precordial leads exhibit non-negligible correlations. After introducing DLE/SFM guidance, the ARE association map becomes more selective, with stronger weights concentrated on a smaller subset of source leads. This suggests that ARE does not merely follow the raw correlation structure of the input signals. Instead, DLE-derived cues reweight the source-to-target association pattern used for latent cross-lead reconstruction.

The guidance map further illustrates the change induced by DLE/SFM guidance. Positive regions indicate enhanced source-to-target associations after guidance, whereas negative regions indicate suppressed associations. The observed pattern shows that DLE/SFM guidance selectively redistributes association weights instead of uniformly increasing all lead relations. This supports the interpretation that DLE provides lead-wise cues for ARE, enabling cue-guided cross-lead consistency modeling in the latent representation space.

\subsection{Observational Study}
We further conducted an observational study on real-world clinical ECG data from an external clinical cohort, without intervening in patient care. This cohort contains population and demographic distributions different from those of the public benchmark datasets, providing a more challenging setting for evaluating representation generalizability. As shown in Table~\ref{tab:observational_study_result}, DLE\&ARE representations improve most metrics across CNN, Transformer, and Mamba classifiers compared with baseline representations. The gains are particularly consistent in accuracy, F1 score, and AUC, indicating that the learned deviation-association representations provide additional transferable information under real-world distribution shifts. Although the absolute performance remains limited due to the heterogeneity of the observational cohort, the consistent relative improvements support the robustness of the proposed feature extraction framework.

\subsection{Ablation Study}

Table~\ref{tab:ablation_study} summarizes the performance impact of individual components in our framework.

\textbf{Pretraining phase.}
Beginning with a random masking strategy and a Transformer backbone, the baseline model achieved B-Acc = 0.289 on CPSC2020 and 0.270 on WGEX. Replacing the Transformer backbone with SSMb under the same hyperparameter setting improved B-Acc to 0.397 and 0.337, respectively, demonstrating the benefits of hierarchical temporal modeling. Adding self-supervised symmetric masking further boosted B-Acc to 0.466 on CPSC2020 and 0.464 on WGEX, while increasing AUROC by over 0.07 on both datasets, highlighting the value of reconstruction-based pretraining. Incorporating weighted rotational partial freezing without DLE and DMb yielded B-Acc = 0.454 on CPSC2020, confirming that selective parameter updates contribute to training stability.

\textbf{Component contributions.}
The variant combining SSMb, SymMask, RPF, DLE, and DMb without ARE achieved B-Acc = 0.473 on CPSC2020 and 0.449 on WGEX, showing cumulative performance gains from these modules. Removing DMb while adding ARE raised B-Acc to 0.500 on CPSC2020 and 0.476 on WGEX, underscoring the role of ARE in inter-lead integration. The full model, incorporating all components, achieved the best results: B-Acc = 0.538, F1 = 0.536, and AUROC = 0.763 on CPSC2020; and B-Acc = 0.513, F1 = 0.471, and AUROC = 0.787 on WGEX. These results confirm the complementary contributions of each module and the overall synergy of the full architecture. 

\section{Conclusion}

In this study, we proposed a new ECG feature extraction framework that integrates multi-lead association embedding and deviation localization embedding. Our method effectively captures deviations from normal ECG morphology while integrating complementary cross-lead contextual information through association reconstruction.  Through extensive experiments across diverse classifiers and datasets, we validated the reliability and generalizability of the framework. Our future work will focus on integrating SOTA ECG-based CVD diagnosis models to further enhance performance and provide more precise, dependable clinical decision support tools.

% \section*{Appendix}

% Appendixes, if needed, appear before the acknowledgment.

% \section*{Acknowledgment}

% The preferred spelling of the word ``acknowledgment'' in American English is without an ``e'' after the ``g.'' Use the singular heading even if you have many acknowledgments. Avoid expressions such as ``One of us (S.B.A.) would like to thank ... .'' Instead, write ``F. A. Author thanks ... .'' In most cases, sponsor and financial support acknowledgments are placed in the unnumbered footnote on the first page, not here.

\bibliographystyle{IEEEtran}
\bibliography{ref}

@article{mckeen2024ecg,
  title   = {{ECG-FM}: An open electrocardiogram foundation model},
  author  = {McKeen, Kaden and Masood, Sameer and Toma, Augustin and Rubin, Barry and Wang, Bo},
  journal = {JAMIA Open},
  volume  = {8},
  number  = {5},
  pages   = {ooaf122},
  year    = {2025},
  doi     = {10.1093/jamiaopen/ooaf122}
}

@article{xiao2023deep,
  title   = {Deep learning-based {ECG} arrhythmia classification: A systematic review},
  author  = {Xiao, Qiao and Lee, Khuan and Mokhtar, Siti Aisah and Ismail, Iskasymar and Pauzi, Ahmad Luqman bin Md and Zhang, Qiuxia and Lim, Poh Ying},
  journal = {Applied Sciences},
  volume  = {13},
  number  = {8},
  pages   = {4964},
  year    = {2023}
}

@article{al2023machine,
  title   = {Machine learning for {ECG} diagnosis and risk stratification of occlusion myocardial infarction},
  author  = {Al-Zaiti, Salah S. and Martin-Gill, Christian and Z{\`e}gre-Hemsey, Jessica K. and Bouzid, Zeineb and Faramand, Ziad and Alrawashdeh, Mohammad O. and Gregg, Richard E. and Helman, Stephanie and Riek, Nathan T. and Kraevsky-Phillips, Karina and others},
  journal = {Nature Medicine},
  volume  = {29},
  number  = {7},
  pages   = {1804--1813},
  year    = {2023}
}

@article{muzammil2024artificial,
  title   = {Artificial intelligence-enhanced electrocardiography for accurate diagnosis and management of cardiovascular diseases},
  author  = {Muzammil, Muhammad Ali and Javid, Saman and Afridi, Azra Khan and Siddineni, Rupini and Shahabi, Mariam and Haseeb, Muhammad and Fariha, FNU and Kumar, Satesh and Zaveri, Sahil and Nashwan, Abdulqadir J.},
  journal = {Journal of Electrocardiology},
  year    = {2024}
}

@article{silvetti2023pivotal,
  title   = {The pivotal role of {ECG} in cardiomyopathies},
  author  = {Silvetti, Elisa and Lanza, Oreste and Romeo, Fabiana and Martino, Annamaria and Fedele, Elisa and Lanzillo, Chiara and Crescenzi, Cinzia and Fanisio, Francesca and Cal{\`o}, Leonardo},
  journal = {Frontiers in Cardiovascular Medicine},
  volume  = {10},
  pages   = {1178163},
  year    = {2023}
}

@article{arslan2024ecg,
  title   = {{ECG} heartbeats classification with dilated convolutional autoencoder},
  author  = {Arslan, Naciye Nur and Ozdemir, Durmus and Temurtas, Hasan},
  journal = {Signal, Image and Video Processing},
  volume  = {18},
  number  = {1},
  pages   = {417--426},
  year    = {2024}
}

@article{gunduz2023atrial,
  title   = {Atrial fibrillation classification and detection from {ECG} recordings},
  author  = {G{\"u}nd{\"u}z, Ali Fatih and Talu, Muhammed Fatih},
  journal = {Biomedical Signal Processing and Control},
  volume  = {82},
  pages   = {104531},
  year    = {2023}
}

@inproceedings{izci2018arrhythmia,
  title     = {Arrhythmia detection on {ECG} signals by using empirical mode decomposition},
  author    = {Izci, Elif and Ozdemir, Mehmet Akif and Sadighzadeh, Reza and Akan, Aydin},
  booktitle = {Proc. Medical Technologies National Congress (TIPTEKNO)},
  pages     = {1--4},
  year      = {2018}
}

@article{hu2020effective,
  title   = {An effective frequency-domain feature of atrial fibrillation based on time--frequency analysis},
  author  = {Hu, Yusong and Zhao, Yantao and Liu, Jihong and Pang, Jin and Zhang, Chen and Li, Peizhe},
  journal = {BMC Medical Informatics and Decision Making},
  volume  = {20},
  pages   = {1--11},
  year    = {2020}
}

@inproceedings{dai2021attentional,
  title     = {Attentional feature fusion},
  author    = {Dai, Yimian and Gieseke, Fabian and Oehmcke, Stefan and Wu, Yiquan and Barnard, Kobus},
  booktitle = {Proc. IEEE/CVF Winter Conf. Applications of Computer Vision},
  pages     = {3560--3569},
  year      = {2021}
}

@article{shiraishi2023improved,
  title   = {Improved prediction of sudden cardiac death in patients with heart failure through digital processing of electrocardiography},
  author  = {Shiraishi, Yasuyuki and Goto, Shinichi and Niimi, Nozomi and Katsumata, Yoshinori and Goda, Ayumi and Takei, Makoto and Saji, Mike and Sano, Motoaki and Fukuda, Keiichi and Kohno, Takashi and others},
  journal = {Europace},
  volume  = {25},
  number  = {3},
  pages   = {922--930},
  year    = {2023}
}

@article{hughes2023deep,
  title   = {A deep learning-based electrocardiogram risk score for long term cardiovascular death and disease},
  author  = {Hughes, J. Weston and Tooley, James and Torres Soto, Jessica and Ostropolets, Anna and Poterucha, Tim and Christensen, Matthew Kai and Yuan, Neal and Ehlert, Ben and Kaur, Dhamanpreet and Kang, Guson and others},
  journal = {NPJ Digital Medicine},
  volume  = {6},
  number  = {1},
  pages   = {169},
  year    = {2023}
}

@article{kavya2015feature,
  title   = {Feature extraction technique for robust and fast visual tracking: A typical review},
  author  = {Kavya, R. and others},
  journal = {International Journal of Emerging Engineering Research and Technology},
  volume  = {3},
  number  = {1},
  pages   = {98--104},
  year    = {2015}
}

@book{gray2012fourier,
  title     = {Fourier Transforms: An Introduction for Engineers},
  author    = {Gray, Robert M. and Goodman, Joseph W.},
  volume    = {322},
  year      = {2012},
  publisher = {Springer Science \& Business Media}
}

@incollection{kurita2021principal,
  title     = {Principal component analysis ({PCA})},
  author    = {Kurita, Takio},
  booktitle = {Computer Vision: A Reference Guide},
  pages     = {1013--1016},
  year      = {2021},
  publisher = {Springer}
}

@article{gholami2012sparse,
  title = {Sparse time--frequency decomposition and some applications},
  author = {Gholami, Ali},
  journal = {IEEE Transactions on Geoscience and Remote Sensing},
  volume = {51},
  number = {6},
  pages = {3598--3604},
  year = {2013},
  doi = {10.1109/TGRS.2012.2220144}
}

@incollection{sadouk2019cnn,
  title = {{CNN} approaches for time series classification},
  author = {Sadouk, Lamyaa},
  booktitle = {Time Series Analysis - Data, Methods, and Applications},
  editor = {Ngan, Chun-Kit},
  publisher = {IntechOpen},
  chapter = {4},
  year = {2019},
  doi = {10.5772/intechopen.81170}
}

@inproceedings{jin2020prediction,
  title     = {Prediction for time series with {CNN} and {LSTM}},
  author    = {Jin, Xuebo and Yu, Xinghong and Wang, Xiaoyi and Bai, Yuting and Su, Tingli and Kong, Jianlei},
  booktitle = {Proc. 11th Int. Conf. Modelling, Identification and Control},
  pages     = {631--641},
  year      = {2020}
}

@article{wang2022ngcu,
  title   = {{NGCU}: A new {RNN} model for time-series data prediction},
  author  = {Wang, Jingyang and Li, Xiaolei and Li, Jiazheng and Sun, Qiuhong and Wang, Haiyao},
  journal = {Big Data Research},
  volume  = {27},
  pages   = {100296},
  year    = {2022}
}

@article{khaldi2023best,
  title   = {What is the best {RNN}-cell structure to forecast each time series behavior?},
  author  = {Khaldi, Rohaifa and El Afia, Abdellatif and Chiheb, Raddouane and Tabik, Siham},
  journal = {Expert Systems with Applications},
  volume  = {215},
  pages   = {119140},
  year    = {2023}
}

@article{liu2024zero,
  title   = {Zero-shot {ECG} classification with multimodal learning and test-time clinical knowledge enhancement},
  author  = {Liu, Che and Wan, Zhongwei and Ouyang, Cheng and Shah, Anand and Bai, Wenjia and Arcucci, Rossella},
  journal = {arXiv preprint arXiv:2403.06659},
  year    = {2024}
}

@book{alpaydin2020introduction,
  title = {Introduction to Machine Learning},
  author = {Alpaydin, Ethem},
  edition = {4th},
  publisher = {MIT Press},
  year = {2020},
  isbn = {9780262043793}
}

@article{liu2022large,
  title   = {A large-scale multi-label {12-lead} electrocardiogram database with standardized diagnostic statements},
  author  = {Liu, Hui and Chen, Dan and Chen, Da and Zhang, Xiyu and Li, Huijie and Bian, Lipan and Shu, Minglei and Wang, Yinglong},
  journal = {Scientific Data},
  volume  = {9},
  pages   = {272},
  year    = {2022},
  doi     = {10.1038/s41597-022-01403-5}
}

@article{alday2020classification,
  title = {Classification of {12-lead ECGs}: The {PhysioNet/Computing in Cardiology Challenge 2020}},
  author = {Perez Alday, Erick Andres and Gu, Annie and Shah, Amit J. and Robichaux, Chad and Wong, An-Kwok Ian and Liu, Chengyu and Liu, Feifei and Bahrami Rad, Ali and Elola, Andoni and Seyedi, Salman and Li, Qiao and Sharma, Ashish and Clifford, Gari D. and Reyna, Matthew A.},
  journal = {Physiological Measurement},
  volume = {41},
  number = {12},
  pages = {124003},
  year = {2020},
  doi = {10.1088/1361-6579/abc960}
}

@article{wang2023adversarial,
  title = {Adversarial spatiotemporal contrastive learning for electrocardiogram signals},
  author = {Wang, Ning and Feng, Panpan and Ge, Zhaoyang and Zhou, Yanjie and Zhou, Bing and Wang, Zongmin},
  journal = {IEEE Transactions on Neural Networks and Learning Systems},
  volume = {35},
  number = {10},
  pages = {13845--13859},
  year = {2024},
  doi = {10.1109/TNNLS.2023.3272153}
}

@article{lai2023practical,
  title   = {Practical intelligent diagnostic algorithm for wearable {12-lead ECG} via self-supervised learning on large-scale dataset},
  author  = {Lai, Jiewei and Tan, Huixin and Wang, Jinliang and Ji, Lei and Guo, Jun and Han, Baoshi and Shi, Yajun and Feng, Qianjin and Yang, Wei},
  journal = {Nature Communications},
  volume  = {14},
  number  = {1},
  pages   = {3741},
  year    = {2023}
}

@inproceedings{zhou2019beatgan,
  title     = {{BeatGAN}: Anomalous rhythm detection using adversarially generated time series},
  author    = {Zhou, Bin and Liu, Shenghua and Hooi, Bryan and Cheng, Xueqi and Ye, Jing},
  booktitle = {Proc. Int. Joint Conf. Artificial Intelligence},
  pages     = {4433--4439},
  year      = {2019}
}

@article{jiang2024self,
  title   = {Self-supervised anomaly detection pretraining enhances long-tail {ECG} diagnosis},
  author  = {Jiang, Aofan and Huang, Chaoqin and Cao, Qing and Xu, Yuchen and Zeng, Zi and Chen, Kang and Zhang, Ya and Wang, Yanfeng},
  journal = {arXiv preprint arXiv:2408.17154},
  year    = {2024}
}

@misc{zheng2022large,
  title = {A large scale {12-lead} electrocardiogram database for arrhythmia study},
  author = {Zheng, Jianwei and Guo, Hangyuan and Chu, Huimin},
  howpublished = {PhysioNet},
  year = {2022},
  note = {Version 1.0.0},
  doi = {10.13026/wgex-er52}
}

@article{wagner2020ptb,
  title   = {{PTB-XL}, a large publicly available electrocardiography dataset},
  author  = {Wagner, P. and Strodthoff, N. and Bousseljot, R. D. and others},
  journal = {Scientific Data},
  volume  = {7},
  number  = {1},
  pages   = {154},
  year    = {2020},
  doi     = {10.1038/s41597-020-0495-6}
}

@article{avanzato2020automatic,
  title   = {Automatic {ECG} diagnosis using convolutional neural network},
  author  = {Avanzato, R. and Beritelli, F.},
  journal = {Electronics},
  volume  = {9},
  number  = {6},
  pages   = {951},
  year    = {2020}
}

@article{nguyen2021deep,
  title   = {Deep learning framework with {ECG} feature-based kernels for heart disease classification},
  author  = {Nguyen, T. N. and Nguyen, T. H.},
  journal = {Elektronika ir Elektrotechnika},
  volume  = {27},
  number  = {1},
  pages   = {48--59},
  year    = {2021}
}

@article{karimian2017highly,
  title = {Highly reliable key generation from electrocardiogram ({ECG})},
  author = {Karimian, Nima and Guo, Zimu and Tehranipoor, Mark and Forte, Domenic},
  journal = {IEEE Transactions on Biomedical Engineering},
  volume = {64},
  number = {6},
  pages = {1400--1411},
  year = {2017},
  doi = {10.1109/TBME.2016.2607020}
}

@article{chazal2004automatic,
  title   = {Automatic classification of heartbeats using {ECG} morphology and heartbeat interval features},
  author  = {De Chazal, P. and O'Dwyer, M. and Reilly, R. B.},
  journal = {IEEE Transactions on Biomedical Engineering},
  volume  = {51},
  number  = {7},
  pages   = {1196--1206},
  year    = {2004}
}

@article{oquab2023dinov2,
  title   = {{DINOv2}: Learning robust visual features without supervision},
  author = {Oquab, Maxime and Darcet, Timoth{\'e}e and Moutakanni, Th{\'e}o and Vo, Huy and Szafraniec, Marc and Khalidov, Vasil and Fernandez, Pierre and Haziza, Daniel and Massa, Francisco and El-Nouby, Alaaeldin and Assran, Mahmoud and Ballas, Nicolas and Galuba, Wojciech and Howes, Russell and Huang, Po-Yao and Li, Shang-Wen and Misra, Ishan and Rabbat, Michael and Sharma, Vasu and Synnaeve, Gabriel and Xu, Hu and Jegou, Herv{\'e} and Mairal, Julien and Labatut, Patrick and Joulin, Armand and Bojanowski, Piotr},
  journal = {arXiv preprint arXiv:2304.07193},
  year    = {2023},
  note    = {Version 1},
  doi     = {10.48550/arXiv.2304.07193}
}

@article{das2023decoder,
  title   = {A decoder-only foundation model for time-series forecasting},
  author  = {Das, Abhimanyu and Kong, Weihao and Sen, Rajat and Zhou, Yichen},
  journal = {arXiv preprint arXiv:2310.10688},
  year    = {2023}
}

@article{woo2024unified,
  title   = {Unified training of universal time series forecasting transformers},
  author  = {Woo, Gerald and Liu, Chenghao and Kumar, Akshat and Xiong, Caiming and Savarese, Silvio and Sahoo, Doyen},
  journal = {arXiv preprint arXiv:2402.02592},
  year    = {2024}
}

@article{gu2021efficiently,
  title   = {Efficiently modeling long sequences with structured state spaces},
  author  = {Gu, Albert and Goel, Karan and R{\'e}, Christopher},
  journal = {arXiv preprint arXiv:2111.00396},
  year    = {2021}
}

@article{smith2022simplified,
  title   = {Simplified state space layers for sequence modeling},
  author  = {Smith, Jimmy T. H. and Warrington, Andrew and Linderman, Scott W.},
  journal = {arXiv preprint arXiv:2208.04933},
  year    = {2022}
}

@article{ansari2024chronos,
  title   = {{Chronos}: Learning the language of time series},
  author  = {Ansari, Abdul Fatir and Stella, Lorenzo and Turkmen, Caner and Zhang, Xiyuan and Mercado, Pedro and Shen, Huibin and Shchur, Oleksandr and Rangapuram, Syama Sundar and Pineda Arango, Sebastian and Kapoor, Shubham and Zschiegner, Jasper and Maddix, Danielle C. and Wang, Hao and Mahoney, Michael W. and Torkkola, Kari and Wilson, Andrew Gordon and Bohlke-Schneider, Michael and Wang, Yuyang},
  journal = {arXiv preprint arXiv:2403.07815},
  year    = {2024},
  note    = {Version 3}
}

@article{gu2023mamba,
  title   = {{Mamba}: Linear-time sequence modeling with selective state spaces},
  author  = {Gu, Albert and Dao, Tri},
  journal = {arXiv preprint arXiv:2312.00752},
  year    = {2023}
}

@inproceedings{dao2024mamba2,
  title = {Transformers are {SSM}s: Generalized models and efficient algorithms through structured state space duality},
  author = {Dao, Tri and Gu, Albert},
  booktitle = {Proceedings of the 41st International Conference on Machine Learning},
  series = {Proceedings of Machine Learning Research},
  volume = {235},
  pages = {10041--10071},
  year = {2024},
  publisher = {PMLR}
}

@article{zhang2024motion,
  title   = {Motion {Mamba}: Efficient and long sequence motion generation with hierarchical and bidirectional selective {SSM}},
  author  = {Zhang, Zeyu and Liu, Akide and Reid, Ian and Hartley, Richard and Zhuang, Bohan and Tang, Hao},
  journal = {arXiv preprint arXiv:2403.07487},
  year    = {2024}
}

@article{patro2024simba,
  title   = {{SiMBA}: Simplified {Mamba}-based architecture for vision and multivariate time series},
  author  = {Patro, Badri N. and Agneeswaran, Vijay S.},
  journal = {arXiv preprint arXiv:2403.15360},
  year    = {2024}
}

@article{patro2024mamba,
  title   = {{Mamba-360}: Survey of state space models as transformer alternative for long sequence modelling: Methods, applications, and challenges},
  author  = {Patro, Badri Narayana and Agneeswaran, Vijay Srinivas},
  journal = {arXiv preprint arXiv:2404.16112},
  year    = {2024}
}

@article{Li2025ECGFounder,
  title   = {An electrocardiogram foundation model built on over 10 million recordings},
  author  = {Li, Jun and Aguirre, Aaron D. and Junior, Valdery Moura and Jin, Jiarui and Liu, Che and Zhong, Lanhai and Sun, Chenxi and Clifford, Gari and Westover, Brandon and Hong, Shenda},
  journal = {NEJM AI},
  volume  = {2},
  number  = {7},
  year    = {2025},
  month   = jul,
  doi     = {10.1056/AIoa2401033},
  note    = {Accepted Mar. 28, 2025; revised Mar. 20, 2025}
}

@inproceedings{kiyasseh2021clocs,
  title     = {{CLOCS}: Contrastive learning of cardiac signals across space, time, and patients},
  author    = {Kiyasseh, Dani and Zhu, Tingting and Clifton, David A.},
  booktitle = {Proc. Int. Conf. Machine Learning},
  pages     = {5606--5615},
  year      = {2021}
}

@article{Oord2018RepresentationLW,
  title   = {Representation learning with contrastive predictive coding},
  author  = {Oord, A{\"a}ron van den and Li, Yazhe and Vinyals, Oriol},
  journal = {arXiv preprint arXiv:1807.03748},
  year    = {2018}
}

@article{Mehari2021SelfsupervisedRL,
  title   = {Self-supervised representation learning from 12-lead {ECG} data},
  author  = {Mehari, Temesgen and Strodthoff, Nils},
  journal = {Computers in Biology and Medicine},
  volume  = {141},
  pages   = {105114},
  year    = {2022},
  doi     = {10.1016/j.compbiomed.2021.105114}
}

@article{Zhang2023MaeFEMA,
  title   = {{MaeFE}: Masked autoencoders family of electrocardiogram for self-supervised pretraining and transfer learning},
  author  = {Zhang, Huaicheng and Liu, Wenhan and Shi, Jiguang and Chang, Sheng and Wang, Hao and He, Jin and Huang, Qijun},
  journal = {IEEE Transactions on Instrumentation and Measurement},
  volume  = {72},
  pages   = {1--15},
  year    = {2023}
}

@article{Na2024GuidingMR,
  title   = {Guiding masked representation learning to capture spatio-temporal relationship of electrocardiogram},
  author  = {Na, Yeongyeon and Park, Minje and Tae, Yunwon and Joo, Sunghoon},
  journal = {arXiv preprint arXiv:2402.09450},
  year    = {2024}
}

@article{Jing2019SelfSupervisedVF,
  title   = {Self-supervised visual feature learning with deep neural networks: A survey},
  author  = {Jing, Longlong and Tian, Yingli},
  journal = {IEEE Transactions on Pattern Analysis and Machine Intelligence},
  volume  = {43},
  number  = {11},
  pages   = {4037--4058},
  year    = {2021},
  doi     = {10.1109/TPAMI.2020.2992393}
}

@article{Chen2020ASF,
  title   = {A simple framework for contrastive learning of visual representations},
  author  = {Chen, Ting and Kornblith, Simon and Norouzi, Mohammad and Hinton, Geoffrey E.},
  journal = {arXiv preprint arXiv:2002.05709},
  year    = {2020}
}

@inproceedings{He2019MomentumCF,
  title     = {Momentum contrast for unsupervised visual representation learning},
  author    = {He, Kaiming and Fan, Haoqi and Wu, Yuxin and Xie, Saining and Girshick, Ross B.},
  booktitle = {Proc. IEEE/CVF Conf. Computer Vision and Pattern Recognition},
  pages     = {9726--9735},
  year      = {2020}
}

@article{Grill2020BootstrapYO,
  title   = {Bootstrap your own latent: A new approach to self-supervised learning},
  author  = {Grill, Jean-Bastien and Strub, Florian and Altch{\'e}, Florent and Tallec, Corentin and Richemond, Pierre H. and Buchatskaya, Elena and Doersch, Carl and Pires, Bernardo {\'A}vila and Guo, Zhaohan Daniel and Azar, Mohammad Gheshlaghi and Piot, Bilal and Kavukcuoglu, Koray and Munos, R{\'e}mi and Valko, Michal},
  journal = {arXiv preprint arXiv:2006.07733},
  year    = {2020}
}

@article{Liu2021SelfsupervisedEP,
  title   = {Self-supervised {ECG} pre-training},
  author  = {Liu, Han and Zhao, Zhenbo and She, Qiang},
  journal = {Biomedical Signal Processing and Control},
  volume  = {70},
  pages   = {103010},
  year    = {2021},
  doi     = {10.1016/j.bspc.2021.103010}
}

@inproceedings{Gedon2021FirstST,
  title     = {First steps towards self-supervised pretraining of the {12-lead ECG}},
  author    = {Gedon, Daniel and Ribeiro, Ant{\^o}nio H. and Wahlstr{\"o}m, Niklas and Sch{\"o}n, Thomas Bo},
  booktitle = {Proc. Computing in Cardiology},
  volume    = {48},
  pages     = {1--4},
  year      = {2021}
}

@misc{gow2023mimicivecg,
  title = {{MIMIC-IV-ECG}: Diagnostic Electrocardiogram Matched Subset},
  author = {Gow, Brian and Pollard, Tom and Nathanson, Larry A. and Johnson, Alistair and Moody, Benjamin and Fernandes, Chrystinne and Greenbaum, Nathaniel and Waks, Jonathan W. and Eslami, Parastou and Carbonati, Tanner and Chaudhari, Ashish and Herbst, Elizabeth and Moukheiber, Dana and Berkowitz, Seth and Mark, Roger and Horng, Steven},
  howpublished = {PhysioNet},
  year = {2023},
  note = {Version 1.0},
  doi = {10.13026/4nqg-sb35}
}

\end{document}

\title{Supplementary Material for Deviation-Aware Pretraining for Cardiovascular Disease Detection from Electrocardiograms}

\author{Siyuan Yan, Haibo Zheng, Feng Lu, Wei Li}

\maketitle

\appendices

\section{Two-Dimensional Projection of DLE and ARE Embeddings}
\label{app:featuremap}

\begin{figure*}[!t]
\centering
\includegraphics[width=0.90\textwidth]{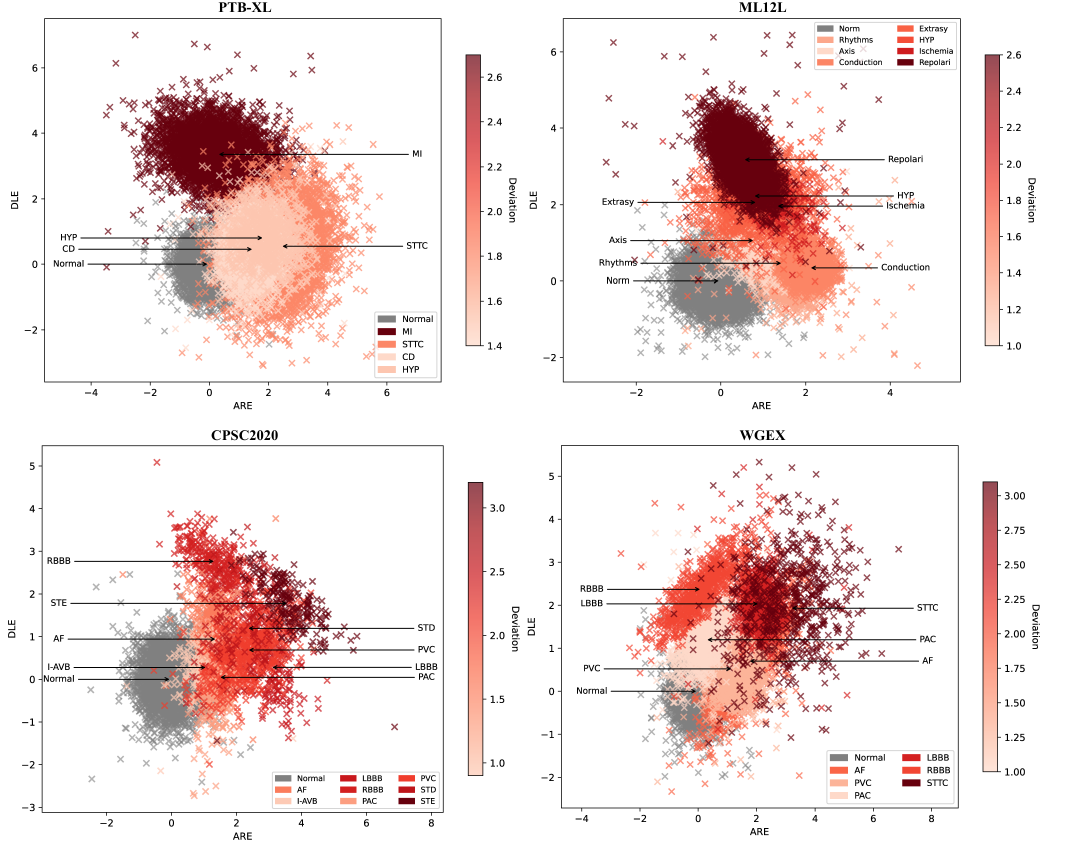}
\caption{Detailed two-dimensional projection of PTB-XL, ML12L, CPSC2020, and WGEX datasets under the DLE (Deviation Localization Embedding) and ARE (Association Reconstruction Embedding) feature embeddings. The x-axis represents the ARE mapping, and the y-axis represents the DLE mapping.}
\label{fig:featureMapDetailed}
\end{figure*}

Fig.~\ref{fig:featureMapDetailed} shows the following observations:

\begin{itemize}
  \item \textbf{PTB-XL:} DLE in Fig.~\ref{fig:featureMapDetailed}(a) forms clear clusters for Myocardial Infarction (MI) and Normal cases but overlaps ST--T Changes (STTC) and Conduction Disorders (CD), reflecting similarities in local waveform deviations. ARE in Fig.~\ref{fig:featureMapDetailed}(e) tightens class boundaries by diffusing subtle MI and CD anomalies across leads, improving class separability.

  \item \textbf{ML12L:} DLE in Fig.~\ref{fig:featureMapDetailed}(b) effectively isolates Hypertrophy (HYP) and Repolarization (Repolari) classes via localized amplitude deviations, but mixes multi-label groups like Rhythms and Conduction. ARE in Fig.~\ref{fig:featureMapDetailed}(f) resolves these by capturing cross-lead associations, yielding more compact, distinct clusters for complex multi-label categories.

  \item \textbf{CPSC2020:} In DLE, Fig.~\ref{fig:featureMapDetailed}(c), atrial fibrillation (AF) and normal samples separate well, yet PVC and PAC remain entangled. ARE in Fig.~\ref{fig:featureMapDetailed}(g) amplifies inter-lead PVC patterns, disentangling PAC and reducing false overlaps with Normal.

  \item \textbf{WGEX:} DLE in Fig.~\ref{fig:featureMapDetailed}(d) highlights PVC and AF anomalies strongly but under-separates RBBB and LBBB, due to subtle morphological differences. ARE in Fig.~\ref{fig:featureMapDetailed}(h) balances this by integrating minor deviations across leads, improving discrimination among bundle-branch blocks.
\end{itemize}

These projections confirm that DLE excels at localizing beat-level deviations, while ARE captures cross-lead anomaly patterns. Their combination underpins the superior classification performance observed in our experiments.

\begin{table*}[!t]
  \centering
  \caption{Overview of the validation datasets and observational data.}
  \label{tab:validation_overview}
  \scriptsize
  \setlength{\tabcolsep}{3pt}
  \renewcommand{\arraystretch}{1.05}
  \begin{tabularx}{\textwidth}{lcccX}
    \toprule
    \textbf{Dataset} & \textbf{\#Samples} & \textbf{\#Classes} & \textbf{Sampling Rate} & \textbf{Classes} \\
    \midrule
    PTB-XL~\cite{wagner2020ptb}  
      & 21\,873 & 5 & 500~Hz 
      & Normal, MI, STTC, CD, HYP \\
    ML12L~\cite{liu2022large} 
      & 25\,770 & 8 & 500~Hz 
      & Norm, Rhythms, Axis, Conduction, Extrasy, HYP, Ischemia, Repolari \\
    CPSC2020~\cite{alday2020classification}  
      & 10\,330 & 9 & 500~Hz 
      & Normal, AF, I-AVB, LBBB, RBBB, PAC, PVC, STD, STE \\
    WGEX~\cite{zheng2022large}
      & 12\,600 & 7 & 500~Hz 
      & Normal, AF, PVC, PAC, LBBB, RBBB, STTC \\
    Observational Data
      & 5\,239 & 7 & 500~Hz 
      & I21, I44, I45, I46, I47, I48, I49 \\
    \bottomrule
  \end{tabularx}
\end{table*}

\begin{table}[!t]
  \centering
  \caption{Class-wise sample counts in each external validation dataset.}
  \label{tab:validation_distributions}
  \scriptsize
  \setlength{\tabcolsep}{3pt}
  \renewcommand{\arraystretch}{1.05}
  \begin{tabular}{lcccc}
    \toprule
    \textbf{Class} & \textbf{PTB-XL} & \textbf{ML12L} & \textbf{CPSC2020} & \textbf{WGEX} \\
    \midrule
    Normal     & 5\,038 & 13\,905 & 3\,855 & 5\,000 \\
    MI         & 4\,508 & \textemdash{} & \textemdash{} & \textemdash{} \\
    STTC       & 5\,367 & \textemdash{} & \textemdash{} & 700 \\
    CD         & 4\,381 & \textemdash{} & \textemdash{} & \textemdash{} \\
    HYP        & 1\,079 & 229 & \textemdash{} & \textemdash{} \\
    Rhythms    & \textemdash{} & 5\,430 & \textemdash{} & \textemdash{} \\
    Axis       & \textemdash{} & 612 & \textemdash{} & \textemdash{} \\
    Conduction & \textemdash{} & 2\,565 & \textemdash{} & \textemdash{} \\
    Extrasy    & \textemdash{} & 1\,689 & \textemdash{} & \textemdash{} \\
    Ischemia   & \textemdash{} & 260 & \textemdash{} & \textemdash{} \\
    Repolari   & \textemdash{} & 5\,125 & \textemdash{} & \textemdash{} \\
    AF         & \textemdash{} & \textemdash{} & 758 & 2\,000 \\
    I-AVB      & \textemdash{} & \textemdash{} & 328 & \textemdash{} \\
    LBBB       & \textemdash{} & \textemdash{} & 222 & 1\,000 \\
    RBBB       & \textemdash{} & \textemdash{} & 320 & 900 \\
    PAC        & \textemdash{} & \textemdash{} & 379 & 1\,200 \\
    PVC        & \textemdash{} & \textemdash{} & 640 & 1\,800 \\
    STD        & \textemdash{} & \textemdash{} & 154 & \textemdash{} \\
    STE        & \textemdash{} & \textemdash{} & 221 & \textemdash{} \\
    \bottomrule
  \end{tabular}
\end{table}

\section{Datasets and Experimental Setup}
\label{app:dataset_setup}

\subsection{Datasets}

Our study leverages two large-scale pretraining datasets and several public benchmark datasets for validation and evaluation.

\subsubsection{Pretraining Datasets}

The pretraining dataset was constructed by combining two large sources of 12-lead ECG recordings, each sampled at 500~Hz with a duration of 10~s:

\begin{itemize}
    \item \textbf{MIMIC-IV-ECG}~\cite{gow2023mimicivecg}: We included all available recordings from this version, excluding only samples with missing waveform data. This contributed approximately 800{,}000 recordings.

    \item \textbf{Private Clinical Dataset}: An additional set of approximately 118{,}000 recordings was sourced from a clinical partner under IRB approval. This dataset was collected in a clinical environment with equipment comparable to that used for the public datasets. It represents a general inpatient population and is not enriched for any specific cardiac condition, containing a mix of healthy and abnormal, but not finely categorized, samples.
\end{itemize}

In total, the pretraining corpus contains approximately 918{,}000 recordings.

\subsubsection{External Evaluation Datasets}

To assess generalization and transfer learning performance, we utilized four distinct public datasets as held-out cohorts:

\begin{itemize}
    \item \textbf{PTB-XL}~\cite{wagner2020ptb}: A large-scale clinical dataset from a general population, widely used for benchmarking general ECG interpretation algorithms.

    \item \textbf{ML12L}~\cite{liu2022large}: A large-scale multi-label 12-lead ECG dataset with standardized diagnostic statements, used to evaluate multi-label cardiovascular disease classification.

    \item \textbf{CPSC2020}~\cite{alday2020classification}: A dataset primarily targeted at the detection of atrial fibrillation and other rhythm disorders.

    \item \textbf{WGEX}: Our evaluation set drawn from \emph{A large scale 12-lead electrocardiogram database for arrhythmia study}~\cite{zheng2022large}, referred to as WGEX in this study.
\end{itemize}

\subsubsection{Observational Datasets}

To evaluate the model extraction performance in real-world scenarios, we conducted observational experiments on real data from a clinical partner in another region.

\subsection{Data Preprocessing and Model Input}

A minimalist preprocessing approach was intentionally adopted to encourage the model to learn robust features directly from near-raw signals. The same procedure was applied to all datasets.

\begin{itemize}
    \item \textbf{Signal Integrity Check:} For the MIMIC dataset, only samples with non-empty ECG waveform data were retained. For the pretraining set, basic outlier detection was performed to remove samples exhibiting extreme artifacts likely due to lead disconnection or severe acquisition errors.

    \item \textbf{No Filtering or Normalization:} The raw waveform data were used directly without applying any signal filters, such as band-pass filtering, or amplitude normalization techniques, such as Z-score normalization. Each 10~s recording was treated as a sequence of 5{,}000 time points per lead.

    \item \textbf{No Data Augmentation:} No data augmentation techniques were applied during the pretraining phase.
\end{itemize}

\subsection{Experimental Design and Evaluation Strategy}

Our experimental protocol is designed to first pretrain a powerful feature extractor in a self-supervised manner and then rigorously evaluate its feature quality and downstream performance.

\subsubsection{Pretraining Phase}

The entire pretraining corpus, including MIMIC-IV-ECG and private data, was used for self-supervised model training. The dataset was shuffled at the recording level. During training, we partitioned this set dynamically, using 80\% of the samples for training updates and 20\% for validation in each epoch to monitor training progress and guide hyperparameter adjustments. It is important to note that this split was based on recording IDs, as the primary goal of our self-supervised pretraining is to learn universal waveform representations rather than patient-specific characteristics.

\subsubsection{Evaluation Phase}

The external datasets, including PTB-XL, ML12L, CPSC2020, and WGEX, were used exclusively for evaluation and were never seen by the model during pretraining. We employed two distinct evaluation strategies:

\begin{enumerate}
    \item \textbf{Linear Evaluation for Feature Quality Assessment:} To assess the intrinsic quality of the learned representations, we performed linear evaluation on the PTB-XL and ML12L datasets. The pretrained ScalesMamba2 model was used as a frozen feature extractor. A simple linear classifier was then trained on top of these extracted features to perform the downstream classification task. The performance of this linear head serves as a direct measure of the features' separability and utility.

    \item \textbf{Comparison With State-of-the-Art Models on Downstream Tasks:} To benchmark against other unsupervised or weakly supervised methods, we evaluated our model on the CPSC2020 and WGEX datasets. Similar to linear probing, we extracted features using our frozen pretrained model and attached a lightweight classification head. The end-to-end performance was then compared against published results from other state-of-the-art models on these specific benchmarks.
\end{enumerate}

\begin{table}[!t]
\centering
\caption{Disease abbreviations and full names in the PTB-XL dataset.}
\label{tab:ptbxl_disease_abbr}
\scriptsize
\setlength{\tabcolsep}{6pt}
\renewcommand{\arraystretch}{1.05}
\begin{tabular}{ll}
\toprule
\textbf{Abbreviation} & \textbf{Full Name} \\
\midrule
Norm & Normal \\
CD   & Conduction Disturbance \\
HYP  & Hypertrophy \\
STTC & ST/T Change \\
MI   & Myocardial Infarction \\
\bottomrule
\end{tabular}
\end{table}

\begin{table}[!t]
\centering
\caption{Disease abbreviations and full names in the CPSC2020 dataset.}
\label{tab:cpsc2020_disease_abbr}
\scriptsize
\setlength{\tabcolsep}{6pt}
\renewcommand{\arraystretch}{1.05}
\begin{tabular}{ll}
\toprule
\textbf{Abbreviation} & \textbf{Full Name} \\
\midrule
Normal & Normal \\
AF     & Atrial Fibrillation \\
I-AVB  & First-degree Atrioventricular Block \\
LBBB   & Left Bundle Branch Block \\
RBBB   & Right Bundle Branch Block \\
PAC    & Premature Atrial Contraction \\
PVC    & Premature Ventricular Contraction \\
STD    & ST-segment Depression \\
STE    & ST-segment Elevation \\
\bottomrule
\end{tabular}
\end{table}

\begin{table}[!t]
\centering
\caption{Disease abbreviations and full names in the WGEX dataset.}
\label{tab:wgex_disease_abbr}
\scriptsize
\setlength{\tabcolsep}{6pt}
\renewcommand{\arraystretch}{1.05}
\begin{tabular}{ll}
\toprule
\textbf{Abbreviation} & \textbf{Full Name} \\
\midrule
Normal & Normal \\
AF     & Atrial Fibrillation \\
PVC    & Premature Ventricular Contraction \\
PAC    & Premature Atrial Contraction \\
LBBB   & Left Bundle Branch Block \\
RBBB   & Right Bundle Branch Block \\
STTC   & ST/T Change \\
\bottomrule
\end{tabular}
\end{table}

\begin{table*}[!t]
\centering
\caption{Cardiovascular disease categories and subcategories in the ML12L dataset.}
\label{tab:ml12l_disease_categories}
\scriptsize
\setlength{\tabcolsep}{4pt}
\renewcommand{\arraystretch}{1.05}
\begin{tabularx}{\textwidth}{>{\raggedright\arraybackslash}p{0.28\textwidth}X}
\toprule
\textbf{Category} & \textbf{Subcategory} \\
\midrule
Norm &  \\
\midrule

\multirow{2}{=}{Electric Axis of the Heart}
& Right-axis deviation \\
& Left-axis deviation \\
\midrule

\multirow{17}{=}{Conduction Abnormalities}
& Sinus rhythm \\
& Short PR interval \\
& AV conduction ratio N:D \\
& Prolonged PR interval \\
& Second-degree AV block, Mobitz type I \\
& Second-degree AV block, Mobitz type II \\
& 2:1 AV block \\
& AV block, varying conduction \\
& AV block, advanced (high-grade) \\
& AV block, complete (third-degree) \\
& Left anterior fascicular block \\
& Left posterior fascicular block \\
& Left bundle-branch block \\
& Incomplete right bundle-branch block \\
& Right bundle-branch block \\
& Ventricular preexcitation \\
& Junctional tachycardia \\
\midrule

\multirow{5}{=}{Extrasystolies}
& Atrial premature complex(es) \\
& Atrial premature complexes, nonconducted \\
& Junctional premature complex(es) \\
& Junctional escape complex(es) \\
& Ventricular premature complex(es) \\
\midrule

\multirow{3}{=}{Hypertrophies}
& Left atrial enlargement \\
& Left ventricular hypertrophy \\
& Right ventricular hypertrophy \\
\midrule

\multirow{7}{=}{Ischemia}
& Atrial fibrillation \\
& Atrial flutter \\
& ST deviation \\
& ST deviation with T-wave change \\
& T-wave abnormality \\
& Prolonged QT interval \\
& TU fusion \\
\midrule

\multirow{3}{=}{Repolarization Abnormalities}
& Ventricular hypertrophy ST-T change \\
& Low voltage \\
& Early repolarization \\
\midrule

\multirow{4}{=}{Myocardial Infarction (MI)}
& Anterior MI \\
& Inferior MI \\
& Anteroseptal MI \\
& Extensive anterior MI \\
\bottomrule
\end{tabularx}
\end{table*}

\begin{table}[!t]
\centering
\caption{Disease categories in the observational dataset.}
\label{tab:observational_dataset}
\scriptsize
\setlength{\tabcolsep}{5pt}
\renewcommand{\arraystretch}{1.05}
\begin{tabular}{clc}
\toprule
\textbf{ICD} & \textbf{Full Name} & \textbf{Size} \\
\midrule
I21 & Acute Myocardial Infarction & 1{,}152 \\
I44 & Atrioventricular and left bundle block & 450 \\
I45 & Other conduction disorders & 207 \\
I46 & Cardiac arrest & 44 \\
I47 & Paroxysmal tachycardia & 700 \\
I48 & Atrial fibrillation and flutter & 2{,}271 \\
I49 & Other cardiac arrhythmias & 415 \\
\bottomrule
\end{tabular}
\end{table}

\section{Prior Knowledge About SSMs and Dynamic Sparse Matrices}
\label{app:ssm_sparse_matrix}

\subsection{State-Space Models and Mambas}

State-space models have consistently shown strong predictive power for time-series data~\cite{gu2021efficiently,smith2022simplified}. The recent development of Mamba and its variations has addressed the challenge of incorporating distant contexts as sequence length increases by integrating the attention mechanism of Transformers into a more streamlined state-space model~\cite{gu2023mamba,dao2024mamba2,patro2024mamba,patro2024simba,zhang2024motion}. ECG data, as a classic example of long time-series data, stands to benefit significantly from Mamba2. Furthermore, the deviation localization extracted from ECG signals, which includes a time dimension, can also be classified as long time-series data. In a self-supervised task, Mamba2 can be used to extract the deviation localization embedding of ECG signals and adapt it dynamically during training. Generating this embedding through Mamba2, as opposed to direct numerical calculation, yields a thorough understanding and aids in the overall feature extraction of ECG signals.

\subsection{Dynamic Sparse Matrices and Feature Fusion for ECG}

In the process of association reconstruction for each lead, it is essential to incorporate dynamic sparse matrix and feature fusion technology into the multi-lead association reconstruction embedding process. Studies have demonstrated the effectiveness of sparse representations in capturing ECG features and supporting heart rate and arrhythmia classification~\cite{izci2018arrhythmia,karimian2017highly,chazal2004automatic}. Concurrently, feature fusion techniques can amalgamate signals from multiple leads to extract comprehensive features and precisely reconstruct ECG signals~\cite{dai2021attentional,hu2020effective}.

\section{Detailed Mechanism for Weighted Rotational Partial Freezing Strategy}
\label{app:rpf}

Standard self-supervised training can sometimes lead to the model ``over-optimizing'' for the pretext task, e.g., masked signal reconstruction, at the expense of learning robust and generalizable features. To address this, we introduce a Weighted Rotational Partial Freezing (RPF) strategy. This advanced approach provides fine-grained control over the learning dynamics of different parts of the network, ensuring a balanced focus on both feature extraction and task-specific learning.

\subsection{Methodology of Weighted RPF}

We partition the model's full parameter vector \(\theta\) into \(G\) disjoint blocks, \(\theta = [\theta_1, \dots, \theta_G]\). These blocks are conceptually grouped into two main components:

\begin{itemize}
    \item \textbf{The Feature Extractor \((\mathcal{E})\):} Comprising parameters \(\{\theta_g \mid g \in \mathcal{I}_{\mathcal{E}}\}\) from the deeper, foundational layers of our architecture, e.g., the encoder path of DMb/SSMb. These are intended to learn transferable representations.

    \item \textbf{The Self-Supervision Head \((\mathcal{H})\):} Comprising parameters \(\{\theta_g \mid g \in \mathcal{I}_{\mathcal{H}}\}\) from the final layers responsible for the reconstruction task, e.g., the decoder and output layers.
\end{itemize}

Unlike a uniform cyclic schedule, weighted RPF selects a block \(g\) to unfreeze at each training step \(t\) according to a non-uniform probability distribution \(\mathbf{p} = [p_1, \dots, p_G]\), where \(p_g = \Pr(g(t)=g)\). The parameter update rule remains:
\begin{equation}
\label{eq:rpf_update_weighted}
    \theta^{(t+1)}
    =
    \theta^{(t)}
    -
    \alpha^{(t)}
    M_{g(t)}
    \odot
    \nabla_{\theta}
    \mathcal{L}
    \left(
    \theta^{(t)}
    \right),
\end{equation}
where \(M_{g(t)}\) is the masking matrix that activates only block \(g(t)\).

\subsection{Controlling Learning Dynamics via Update Probabilities}

The core of weighted RPF is the deliberate, uneven allocation of optimization ``budget.'' We assign higher update probabilities to the feature extractor blocks and lower probabilities to the self-supervision head blocks. This is formalized by two principles:

\begin{enumerate}
    \item \textbf{Prioritizing Feature Learning:} We set \(p_i > p_j\) for any extractor block \(\theta_i \in \mathcal{E}\) and any head block \(\theta_j \in \mathcal{H}\). This ensures that, on average, the feature extractor receives more frequent updates.

    \item \textbf{Capping Head Optimization:} We introduce a hyperparameter \(\gamma_{\max} \in (0,1)\) to impose an upper bound on the total probability mass allocated to the self-supervision head:
    \begin{equation}
    \label{eq:rpf_gamma_constraint}
        \sum_{g \in \mathcal{I}_{\mathcal{H}}} p_g \le \gamma_{\max}.
    \end{equation}
    For instance, setting \(\gamma_{\max}=0.50\) guarantees that the reconstruction-specific parts of the model receive at most 25\% of the total parameter updates.
\end{enumerate}

\subsection{Architectural Justification and Benefits}

This weighted strategy provides a principled approach to guiding the learning process with several key benefits:

\begin{itemize}
    \item \textbf{Guided Optimization Focus:} By allocating more updates to the feature extractor \(\mathcal{E}\), W-RPF explicitly directs the model's optimization effort toward learning deep, meaningful representations from the data, rather than superficial patterns sufficient only for the pretext task.

    \item \textbf{Strong Regularization of the Task Head:} Severely limiting the update frequency of the head \(\mathcal{H}\) acts as a powerful form of regularization. It prevents the head from overfitting to the reconstruction task and forces the feature extractor to produce high-quality, informative features that make the reconstruction task easier for a ``lazy'' head.

    \item \textbf{Mitigation of Gradient Interference:} The fundamental benefit of modular updates is retained. By isolating the update of each block, W-RPF prevents large, clean gradients from the reconstruction head from interfering with the smaller, more nuanced gradients required to refine the deep feature extractor, leading to more stable training.

    \item \textbf{Principled and Flexible Control:} W-RPF transforms the training process from a simple uniform cycle into a controllable and targeted strategy. The probability vector \(\mathbf{p}\) and the cap \(\gamma_{\max}\) become meaningful hyperparameters that allow a researcher to precisely balance the trade-off between pretext task performance and the quality of the learned features, adapting the strategy to different datasets and architectural depths.
\end{itemize}

\section{Application of Symmetric Masking to Self-Supervised Feature Learning}
\label{app:symmask_rationale}

We compare two masking strategies for self-supervised reconstruction: \emph{random masking}, where each position is masked independently with probability \(p\), and \emph{symmetric masking}, where two complementary masks cover all positions.

\subsection{Loss Definitions}

Let \(\mathbf{x} \in \mathbb{R}^{T}\) be an ECG segment and \(f(\cdot)\) our reconstruction network. Under random masking \(M \in \{0,1\}^{T}\) with \(M_t \sim \mathrm{Bernoulli}(p)\), the expected reconstruction loss is
\begin{equation}
\mathcal{L}_{\mathrm{rand}}
=
\mathbb{E}_{M}
\left[
\left\|
M \odot
\left(
f\left((\mathbf{1}-M)\odot\mathbf{x}\right)
-
\mathbf{x}
\right)
\right\|^{2}
\right].
\end{equation}

Under symmetric masking, we sample \(M^{(1)} \sim \mathrm{Bernoulli}(p)\) and set \(M^{(2)} = \mathbf{1} - M^{(1)}\). The combined loss is
\begin{align}
\mathcal{L}_{\mathrm{sym}}
=
\mathbb{E}_{M^{(1)}}
\Bigl[
&
\left\|
M^{(1)}
\odot
\left(
f\left(M^{(2)}\odot\mathbf{x}\right)
-
\mathbf{x}
\right)
\right\|^{2}
\nonumber\\
+
&
\left\|
M^{(2)}
\odot
\left(
f\left(M^{(1)}\odot\mathbf{x}\right)
-
\mathbf{x}
\right)
\right\|^{2}
\Bigr].
\end{align}

\subsection{Unbiased Gradient Estimate}

Denote the per-position gradient for coordinate \(t\) as \(g_t(\mathbf{x})\). Under random masking,
\begin{equation}
\mathbb{E}
\left[
M_t g_t(\mathbf{x})
\right]
=
p g_t(\mathbf{x}),
\end{equation}
so the expected update only covers a fraction \(p\) of the full gradient. In contrast, symmetric masking yields
\begin{equation}
\mathbb{E}
\left[
M^{(1)}_t g_t^{(2)} + M^{(2)}_t g_t^{(1)}
\right]
=
p g_t^{(2)}
+
(1-p) g_t^{(1)},
\end{equation}
where \(g_t^{(i)}\) is the gradient from mask \(M^{(i)}\). If \(f\) is sufficiently expressive, both gradients approximate the true gradient \(g_t\), so
\begin{equation}
\mathbb{E}
\left[
\nabla_t \mathcal{L}_{\mathrm{sym}}
\right]
\approx
g_t(\mathbf{x}),
\end{equation}
i.e., symmetric masking recovers an \emph{unbiased} estimate of the full-signal gradient, unlike random masking, which scales it by \(p\).

\subsection{Variance Reduction}

Let \(\sigma^2\) be the variance of the full gradient at coordinate \(t\). For random masking,
\begin{equation}
\mathrm{Var}
\left(
M_t g_t
\right)
=
p(1-p)g_t^2
<
p\sigma^2.
\end{equation}

Under symmetric masking, the variance of the sum of two complementary terms is
\begin{align}
&
\mathrm{Var}
\left(
M^{(1)}_t g^{(2)}_t
+
M^{(2)}_t g^{(1)}_t
\right)
\nonumber\\
&=
p(1-p)
\left(
g^{(2)}_t
-
g^{(1)}_t
\right)^2
\lesssim
p(1-p)\sigma^2,
\end{align}
which is at most half that of the random-mask case when \(p \approx 0.5\). Thus, symmetric masking provides a lower-variance gradient estimate.

\subsection{Full-Coverage Property}

By construction, \(M^{(1)}_t + M^{(2)}_t = 1\), so every position is reconstructed exactly once per sample, ensuring no data region is systematically under-trained. Random masking leaves a fraction \(1-p\) of positions never seen in reconstruction for a given \(M\).

Symmetric masking yields an unbiased, lower-variance gradient estimate that covers all positions every iteration, leading to more stable and efficient learning of ECG waveform morphology compared with standard random masking.

\section{Architectural Rationale: Mamba2's Superiority for ECG Anomaly Detection}
\label{app:mamba_rationale}

The analysis of electrocardiogram (ECG) signals necessitates the detection of transient, often low-amplitude, anomalies within a long-duration, quasi-periodic context. An effective architecture must not only identify these localized events but also preserve their fidelity against dilution from the surrounding signal. Here, we provide a rationale for why the Mamba2 architecture, built upon the Selective State Space Model (Selective SSM), is structurally better suited for this task than Transformer, CNN, and traditional RNN models.

\subsection{Mamba2 vs. Transformer: State Compression Over Global Averaging}

The self-attention mechanism in Transformers computes a global, weighted average of the entire sequence. For an input signal \(x_t = s_t + \delta\,\mathbf{1}_{t=t_0}\), where \(\delta\) is a weak, local anomaly, the output is:
\begin{equation}
    o_t =
    \sum_{k=1}^{T}
    \alpha_{t,k} V_k,
    \quad
    \text{where}
    \quad
    \alpha_{t,k}
    =
    \mathrm{softmax}
    \left(
    Q_t^\top K_k
    \right).
\end{equation}
This formulation poses a structural challenge: the anomaly \(\delta\) must ``compete'' for attention against all other \(T-1\) time points. In a long and noisy ECG, this creates a risk of \textbf{attention dilution}, where the anomaly's influence is smoothed out by the global averaging process, even if it receives a higher weight.

Mamba2 circumvents this by replacing global averaging with \textbf{state compression}. It sequentially processes the input, compressing the history into an evolving latent state \(x_\tau\). The core of this process is the Selective SSM update rule:
\begin{equation}
\label{eq:mamba_selective_update}
    x_\tau =
    A_\tau(h_\tau) \cdot x_{\tau-1}
    +
    B_\tau(h_\tau) \cdot h_\tau .
\end{equation}
Crucially, the state transition matrix \(A_\tau\) and input matrix \(B_\tau\) are functions of the current input \(h_\tau\). This \textbf{selective gating} mechanism allows the model to:
\begin{enumerate}
    \item \textbf{Preserve Salient Inputs:} Upon detecting an anomaly in \(h_{\tau_0}\), the model can dynamically increase the norm of the input gate \(B_{\tau_0}\), ensuring the anomaly is robustly injected into the state \(x_{\tau_0}\) without global competition.

    \item \textbf{Memorize Critical Information:} It can simultaneously adjust the memory gate \(A_{\tau_0}\) to be close to identity, effectively ``latching'' the anomaly information in the state and preventing it from being overwritten by subsequent, less relevant signals.
\end{enumerate}

\subsection{Mamba2 vs. CNN: Dynamic, Content-Aware Filtering Over Static Kernels}

A standard CNN applies a fixed, \textbf{time-invariant} kernel \(w\) to extract features:
\begin{equation}
    y_t =
    \sum_{k}
    w_k \cdot x_{t-k}.
\end{equation}
This static template matching is effective for regular, repeating patterns but struggles with the morphological diversity of ECG anomalies. Mamba2's mechanism, as described in \eqref{eq:mamba_selective_update}, is inherently \textbf{time-variant} and \textbf{content-aware}. Its effective ``filter'' is dynamically configured by the input at each time step, allowing it to adaptively capture a wide variety of anomalous shapes. Furthermore, its recurrent nature inherently models long-range temporal dependencies, a task that requires deep CNN stacks or large dilated convolutions.

\subsection{Mamba2 vs. Traditional RNNs: State Expansion Over Information Bottleneck}

While traditional RNNs, e.g., LSTM and GRU, also use a recurrent state, they suffer from an \textbf{information bottleneck}. All features across all input channels must be compressed into a single, fixed-size state vector \(h_t \in \mathbb{R}^{D}\), leading to feature interference:
\begin{equation}
    h_t =
    f(h_{t-1}, x_t),
    \quad
    \text{where}
    \quad
    \dim(h_t)
    \text{ is fixed.}
\end{equation}
Mamba2 overcomes this with \textbf{state expansion}. For each of its \(D\) input channels, or heads, it maintains an independent, higher-dimensional latent state \(x_\tau \in \mathbb{R}^{N}\), where \(N > 1\) is the state expansion factor. This expands the model's total memory capacity to \(\mathcal{O}(D \times N)\), allowing it to track multiple, distinct temporal features in parallel without them corrupting one another. This is critical for ECG analysis, where the model may need to simultaneously monitor baseline wander, heart rate, and QRS morphology. Additionally, Mamba2's formulation allows for efficient, parallelizable training, \(\mathcal{O}(L)\), breaking the sequential computation bottleneck of traditional RNNs.

In summary, Mamba2's architecture provides a powerful and well-suited inductive bias for detecting localized anomalies in long time-series data. By combining \textbf{state compression} to avoid signal dilution, \textbf{selective gating} to dynamically preserve important features, and \textbf{state expansion} to prevent information bottlenecks, it structurally addresses the key limitations of Transformer, CNN, and traditional RNN models for tasks such as ECG analysis.

\section{Architecture of Anomaly Recognition With ScalesMamba2}
\label{app:dmb_ssmb}

\subsection{The ScalesMamba2 Layer: A Multi-Scale Selective SSM}

At the core of our proposed architecture lies the ScalesMamba2 layer, which integrates a Selective State Space Model (Selective SSM) within a multi-scale residual framework. The foundation is Mamba2's SSM, which models a sequence \(\mathbf{u}_t\) through an evolving latent state \(\mathbf{x}_t\):
\begin{equation}
\label{eq:selective_ssm_scalesmamba}
\begin{aligned}
  \mathbf{h}_t &= W_h \mathbf{u}_t + \mathbf{b}_h, \\
  \mathbf{x}_t &= \mathbf{A}_t(\mathbf{h}_t)\mathbf{x}_{t-1}
  + \mathbf{B}_t(\mathbf{h}_t)\mathbf{h}_t, \\
  \mathbf{y}_t &= \mathbf{C}_t(\mathbf{h}_t)\mathbf{x}_t .
\end{aligned}
\end{equation}
Unlike a standard time-varying model, the state transition matrices \(\mathbf{A}_t\) and input matrices \(\mathbf{B}_t\) are dynamically parameterized as functions of the current input \(\mathbf{h}_t\). This \textbf{content-aware selectivity} allows the model to adapt its behavior in real time, for instance, by amplifying the influence of a detected anomaly via \(\mathbf{B}_t\) or extending its memory of that event via \(\mathbf{A}_t\).

The ScalesMamba2 layer wraps this powerful core in a residual, multi-scale pattern, crucial for preserving signal fidelity across different resolutions:
\begin{equation}
\label{eq:scalesmamba_layer}
  \mathrm{ScalesMamba2}(\mathbf{U})
  =
  \mathbf{U}
  +
  \mathrm{Up}
  \left(
  \mathrm{Mamba2}
  \left(
  \mathrm{Down}(\mathbf{U})
  \right)
  \right),
\end{equation}
where \(\mathrm{Down}\) and \(\mathrm{Up}\) handle temporal resolution changes. This structure ensures that the SSM operates on a compressed representation while the residual connection preserves high-frequency details.

\subsection{The DMb and SSMb Blocks: Recurrent U-Net Structures}

We propose two U-Net-like blocks, the Deviation Mamba2 Block (DMb) and the Self-Supervised Mamba2 Block (SSMb), which leverage stacked ScalesMamba2 layers in an encoder-decoder configuration with skip connections.

\subsubsection{Deviation Mamba2 Block}

Designed for refining a pre-computed deviation signal \(\boldsymbol{\sigma}\), the 3-scale DMb follows this structure:

% \begin{equation}
% \begin{aligned}
% \text{Encoder Path:} \quad 
% & \mathbf{e}^{(0)} = \boldsymbol{\sigma}, \\
% & \mathbf{e}^{(i)} = \mathrm{Mamba2}\bigl(\mathrm{Down}(\mathbf{e}^{(i-1)})\bigr), \\
% &i=1,2,3. \\[1ex]
% \text{Decoder Path:} \quad 
% & \mathbf{d}^{(0)} = \mathbf{e}^{(3)}, \\
% & \mathbf{d}^{(j)} = \mathrm{Mamba2}\bigl(\mathrm{Up}(\mathbf{d}^{(j-1)}) + \mathbf{e}^{(3-j)}\bigr), \\
% &j=1,2,3. \\[1ex]
% \text{Output:} \quad 
% & \mathrm{DMb}(\boldsymbol{\sigma}) = \mathbf{d}^{(3)}.
% \end{aligned}
% \end{equation}

\begin{equation}
\label{eq:dmb_structure}
\begin{aligned}
\text{Encoder Path:}\quad
& \mathbf{e}^{(0)} = \boldsymbol{\sigma}, \\
& \mathbf{e}^{(i)} =
\mathrm{Mamba2}
\left(
\mathrm{Down}
\left(
\mathbf{e}^{(i-1)}
\right)
\right), \\
& i = 1,2,3, \\[1ex]
\text{Decoder Path:}\quad
& \mathbf{d}^{(0)} = \mathbf{e}^{(3)}, \\
& \mathbf{d}^{(j)} =
\mathrm{Mamba2}
\left(
\mathrm{Up}
\left(
\mathbf{d}^{(j-1)}
\right)
+
\mathbf{e}^{(3-j)}
\right), \\
& j = 1,2,3, \\[1ex]
\text{Output:}\quad
& \mathrm{DMb}(\boldsymbol{\sigma}) = \mathbf{d}^{(3)}.
\end{aligned}
\end{equation}

This architecture forces the model to capture anomaly ``fingerprints'' at progressively coarser temporal scales in the encoder, while the decoder uses skip connections to fuse this contextual understanding with fine-grained morphological details for precise anomaly reconstruction.

\subsubsection{Self-Supervised Mamba2 Block}

For the more demanding task of masked segment reconstruction from raw signal \(\mathbf{x}\), the deeper 5-scale SSMb is employed:

\begin{equation}
\label{eq:ssmb_structure}
\begin{aligned}
\text{Encoder Path:}\quad
& \mathbf{u}^{(0)} = \mathbf{x}, \\
& \mathbf{u}^{(i)} =
\mathrm{Mamba2}
\left(
\mathrm{Down}
\left(
\mathbf{u}^{(i-1)}
\right)
\right), \\
& i = 1,\dots,5, \\[1ex]
\text{Decoder Path:}\quad
& \mathbf{v}^{(0)} = \mathbf{u}^{(5)}, \\
& \mathbf{v}^{(j)} =
\mathrm{Mamba2}
\left(
\mathrm{Up}
\left(
\mathbf{v}^{(j-1)}
\right)
+
\mathbf{u}^{(5-j)}
\right), \\
& j = 1,\dots,5, \\[1ex]
\text{Output:}\quad
& \mathrm{SSMb}(\mathbf{x}) = \mathbf{v}^{(5)}.
\end{aligned}
\end{equation}

The increased depth expands the model's effective receptive field exponentially, enabling the bottleneck layer to reason about long-range global rhythm and context, which is essential for plausible reconstruction of heavily masked regions.

\subsection{Architectural Design Rationale}

The efficacy of these recurrent U-Net structures for ECG analysis is rooted in the unique synergy between the U-Net topology and the Mamba2 core.

\begin{itemize}
  \item \textbf{Synergistic Multi-scale Processing:} The U-Net's downsampling path creates coarse-grained representations, compelling the model to learn context over longer time windows. While this can attenuate the amplitude of sharp, transient anomalies, Mamba2's \textbf{selective SSM} excels at detecting and latching onto these weakened but contextually significant signals at low resolutions. This is a task where static CNN kernels might fail. The skip connections then re-inject high-fidelity, high-frequency information to the decoder, allowing the Mamba2 layers there to perform precise, context-informed reconstruction.

  \item \textbf{Modeling of Non-Stationary Dynamics:} ECG signals are inherently non-stationary, e.g., changing heart rates and ST-segment shifts. Mamba2's \textbf{content-aware state transitions}, as defined in \eqref{eq:selective_ssm_scalesmamba}, provide the ideal mechanism for adaptively modeling these dynamics at each time scale, a feat beyond the reach of fixed-kernel CNNs.

  \item \textbf{Hierarchical Context for Reconstruction:} The deep, 5-scale SSMb captures a rich hierarchy of features. At the deepest level, i.e., the bottleneck, Mamba2's long-range memory capacity is leveraged to understand global cardiac rhythm. At shallower levels, it models local P-QRS-T waveform dynamics. The decoder then masterfully integrates this hierarchical information, guided by skip connections, to generate reconstructions that are both globally plausible and morphologically precise.

  \item \textbf{Specialization through Depth Trade-off:} The choice of depth is a deliberate trade-off. The 3-scale DMb is optimized for refining a deviation signal, where the primary task is to enhance localized, pre-identified anomalies. A moderate receptive field is sufficient. In contrast, the 5-scale SSMb, with its vast receptive field \((\propto 2^5)\), is designed for the more complex self-supervision task of reconstructing raw signals from large gaps, which demands a deep understanding of global context.
\end{itemize}

\section{Shape-Based Barycenter MDTW Clustering With T-Wave-Based Initialization}
\label{app:mdtw_twave}

Here we describe in detail a variant of our shape-based barycenter MDTW clustering that first aligns cluster centers to characteristic T-wave positions before performing multi-dimensional DTW (MDTW)-based clustering and barycenter averaging.

\subsection{Preprocessing and T-Wave Center Selection}
\label{app:twave_preproc}

Let \(\{\widetilde{n}_r\}_{r=1}^{m} \subset \mathbb{R}^{T \times L}\) be the preprocessed, normalized ECG segments, each of length \(T\) in time and \(L\) leads.

\begin{enumerate}
  \item \textbf{T-wave detection:} For each segment \(\widetilde{n}_r\), detect its T-wave peak time \(t_r\).

  \item \textbf{Sort and cycle computation:} Sort these times:
    \begin{equation}
    \begin{aligned}
      t_{(1)} &\le t_{(2)} \le \cdots \le t_{(m)}, \\
      T_b
      &=
      \frac{1}{m-1}
      \sum_{r=1}^{m-1}
      \left(
      t_{(r+1)} - t_{(r)}
      \right).
    \end{aligned}
    \end{equation}

  \item \textbf{Reference time points:} Choose a target cluster count \(C\). Define
    \begin{equation}
      \tau_c
      =
      t_{(1)}
      +
      (c-1)
      \frac{T_b}{C},
      \quad
      c = 1,\dots,C.
    \end{equation}

  \item \textbf{Initial centers:} For each \(c\), pick
    \begin{equation}
    \begin{aligned}
      r_c
      &=
      \arg\min_{1 \le r \le m}
      \left|
      t_r - \tau_c
      \right|, \\
      \mu_c^{(0)}
      &=
      \widetilde{n}_{r_c}.
    \end{aligned}
    \end{equation}
\end{enumerate}

\subsection{Multi-Dimensional DTW Distance}
\label{app:mdtw_distance}

For any two segments \(X,Y \in \mathbb{R}^{T \times L}\), define the per-point cost
\begin{equation}
  d
  \left(
  X(\alpha),
  Y(\beta)
  \right)
  =
  \sum_{\ell=1}^{L}
  \left|
  X_{\ell}(\alpha)
  -
  Y_{\ell}(\beta)
  \right|^{2},
\end{equation}
and let \(\mathcal{P}\) be the set of all monotonic warping paths from \((1,1)\) to \((T,T)\). Then
\begin{equation}
\label{eq:mdtw_distance}
  D_{\mathrm{MDTW}}(X,Y)
  =
  \min_{\pi \in \mathcal{P}}
  \sum_{(\alpha,\beta) \in \pi}
  d
  \left(
  X(\alpha),
  Y(\beta)
  \right).
\end{equation}

\subsection{T-Wave-Initialized MDTW Clustering Algorithm}
\label{app:dbi_update_twave_detailed}

Given the T-wave-based initial centers \(\{\mu_c^{(0)}\}_{c=1}^{C}\), we iterate for \(q=0,\dots,Q-1\) as follows:

\begin{enumerate}
  \item \textbf{Assignment, Cluster E-step:}
    \begin{equation}
      \mathcal{C}_c^{(q)}
      =
      \left\{
      r :
      c =
      \arg\min_{j=1,\dots,C}
      D_{\mathrm{MDTW}}
      \left(
      \widetilde{n}_r,
      \mu_j^{(q)}
      \right)
      \right\}.
    \end{equation}

  \item \textbf{Alignment, MDTW paths:}
    For each cluster \(c\) and each \(r \in \mathcal{C}_c^{(q)}\), compute
    \begin{equation}
      \pi_{r,c}
      =
      \arg\min_{\pi \in \mathcal{P}}
      \sum_{(\alpha,\beta) \in \pi}
      d
      \left(
      \mu_c^{(q)}(\alpha),
      \widetilde{n}_r(\beta)
      \right).
    \end{equation}

  \item \textbf{Barycenter Accumulation, M-step:}
    For each cluster \(c\), each time index \(\alpha\), and lead \(\ell\):
    \begin{equation}
    \begin{aligned}
      \mathrm{Acc}_c(\alpha,\ell)
      &=
      \sum_{r \in \mathcal{C}_c^{(q)}}
      \sum_{(\alpha,\beta) \in \pi_{r,c}}
      \widetilde{n}_{r,\ell}(\beta), \\
      \mathrm{Cnt}_c(\alpha)
      &=
      \sum_{r \in \mathcal{C}_c^{(q)}}
      \left|
      \left\{
      \beta :
      (\alpha,\beta) \in \pi_{r,c}
      \right\}
      \right|.
    \end{aligned}
    \end{equation}

  \item \textbf{Update Barycenter:}
    \begin{equation}
    \label{eq:dbi_update_twave_detailed}
    \begin{aligned}
    \mu_{c,\ell}^{(q+1)}(\alpha)
    &=
    \begin{cases}
    \displaystyle
    \frac{\mathrm{Acc}_c(\alpha,\ell)}
    {\mathrm{Cnt}_c(\alpha)},
    & \mathrm{Cnt}_c(\alpha)>0, \\[6pt]
    \mu_{c,\ell}^{(q)}(\alpha),
    & \text{otherwise,}
    \end{cases}
    \\
    &\quad \ell=1,\dots,L.
    \end{aligned}
    \end{equation}
\end{enumerate}

Repeat until convergence or \(q=Q\). The final set \(\{\mu_c^{(Q)}\}\) are the cluster barycenters, each initialized at a representative T-wave position and refined via MDTW barycenter averaging.

\subsection{Comparison Between Selected Sample and MDTW-Aligned Codebook Sample}
\label{app:mdtw_fig}

\begin{figure*}[!t]
\centering
\includegraphics[width=0.70\textwidth]{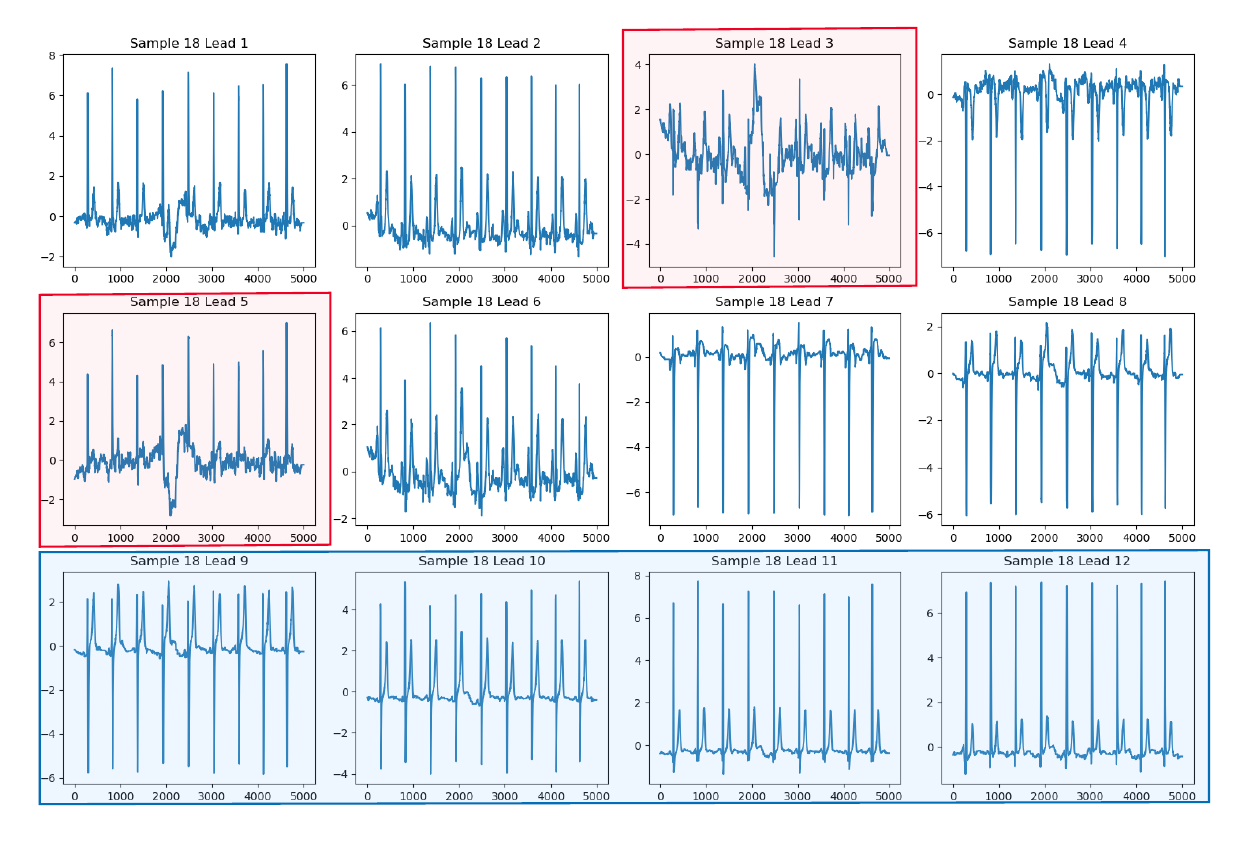}
\caption{One sample in the NormECG cluster selected as the central point. The red box shows leads with slightly abnormal signal acquisition due to factors unrelated to cardiovascular disease, while the blue or unmarked regions indicate the standard lead form of a normal ECG.}
\label{fig:selectedSample}
\end{figure*}

\begin{figure*}[!t]
\centering
\includegraphics[width=0.70\textwidth]{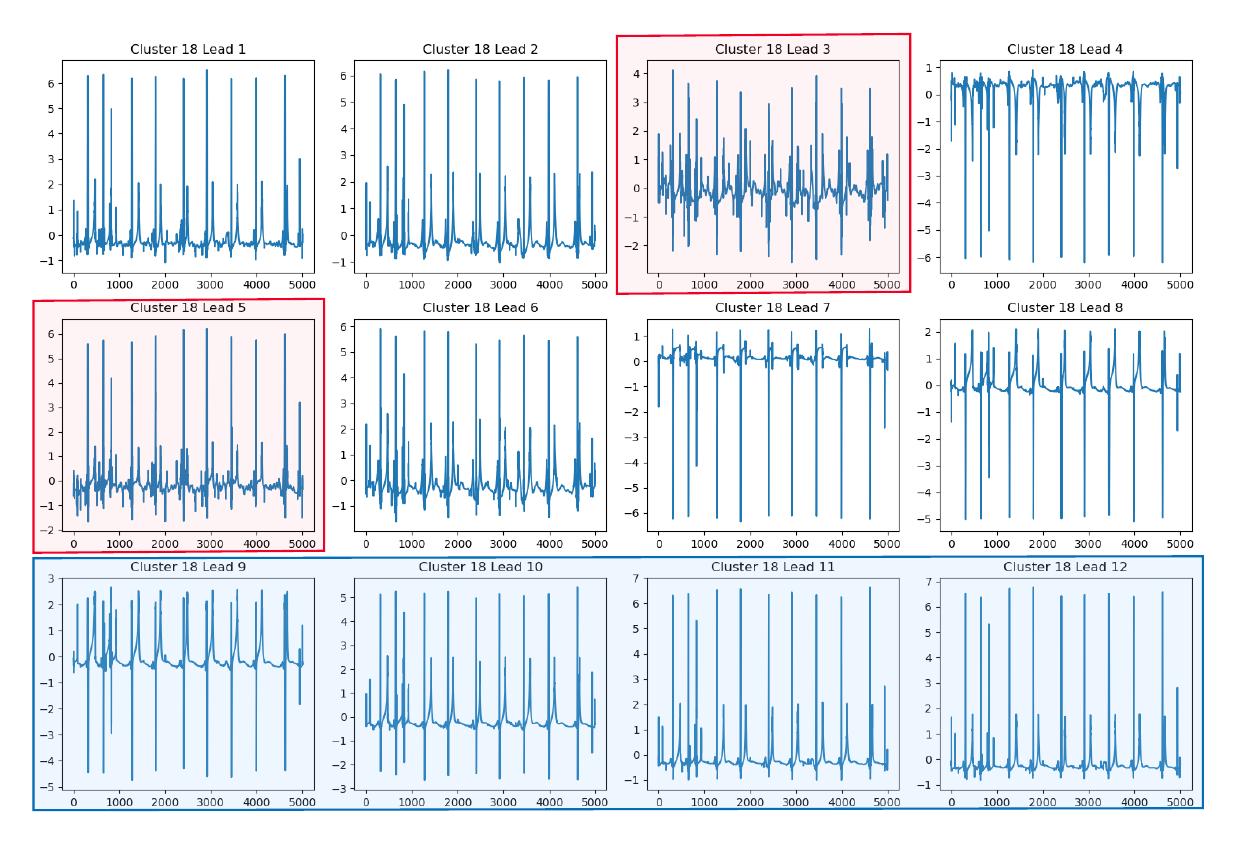}
\caption{An MDTW-aligned sample in the NormECG Codebook corresponding to the selected sample in Fig.~\ref{fig:selectedSample}. The red box again highlights the leads that previously had slightly abnormal signal acquisition, which are now more consistent with typical normal ECG waveforms. The remaining leads, shown in blue or unmarked regions, are mostly unchanged and retain the characteristic shape. Such MDTW fusion helps mitigate external artifacts and preserve essential waveform patterns for subsequent R-peak detection and anomaly assessment.}
\label{fig:MDTWSample}
\end{figure*}

Below we compare: 1) the raw ECG segment chosen as the cluster center based on T-wave-based initialization, as shown in Fig.~\ref{fig:selectedSample}; and 2) the corresponding MDTW-aligned codebook sample after barycenter fusion, as shown in Fig.~\ref{fig:MDTWSample}. The fusion process yields a more standardized morphology, particularly correcting non-pathological fluctuations in leads 3 and 5.

\subsection{Soft-DTW Barycenter}
\label{app:method2_softdtw}

\textit{Soft-DTW} is a differentiable relaxation of the classic DTW objective, controlled by a smoothing parameter \(\lambda>0\). Suppose we have the same cluster \(\{\widetilde{n}_1,\ldots,\widetilde{n}_m\}\) as before. For a candidate barycenter \(\mu\), the Soft-DTW distance to a segment \(\widetilde{n}_r\) can be viewed as:
\begin{equation}
\label{eq:softdtw_dist}
   \mathcal{D}_{\mathrm{soft}}(\mu,\widetilde{n}_r)
   =
   \min_{\pi\in \mathcal{P}}
   \sum_{(\alpha,\beta)\in \pi}
   \left\|
   \mu(\alpha)-\widetilde{n}_r(\beta)
   \right\|^2
   +
   \lambda \Phi(\pi),
\end{equation}
where \(\Phi(\pi)\) is a smooth penalty approximating the discrete warping cost. Soft-DTW then defines the barycenter \(\mu\) by minimizing the aggregate cost over all segments:
\begin{equation}
\label{eq:softdtw_barycenter}
   \mu_{\mathrm{SDTW}}
   =
   \arg\min_{\mu}
   \sum_{r=1}^{m}
   \mathcal{D}_{\mathrm{soft}}(\mu,\widetilde{n}_r).
\end{equation}

Because \(\mathcal{D}_{\mathrm{soft}}\) is differentiable in \(\mu\), we can solve \eqref{eq:softdtw_barycenter} via gradient-based optimization, e.g.,
\begin{equation}
\label{eq:softdtw_update}
   \mu
   \leftarrow
   \mu
   -
   \alpha
   \nabla_{\mu}
   \left(
   \sum_{r=1}^{m}
   \mathcal{D}_{\mathrm{soft}}(\mu,\widetilde{n}_r)
   \right),
\end{equation}
where \(\alpha\) is the learning rate. In practice, if the implementation supports multi-dimensional Soft-DTW, one can directly handle \((T\times L)\) data within each iteration. The final solution is \(\mu_{c}^{\mathrm{SDTW}}\), serving as the smooth DTW-based barycenter for cluster~\(c\).

\subsection{NormECG Codebook Construction}
\label{app:normecg_codebook}

Using either DBI in Appendix~\ref{app:dbi_update_twave_detailed} or Soft-DTW in Appendix~\ref{app:method2_softdtw}, we obtain a barycenter \(\mu_c\) for each cluster \(c=1,\dots,C\). The set of these barycenters forms our \textbf{NormECG Codebook}:
\begin{equation}
\label{eq:normecg_codebook}
   \mathcal{B}_{\mathrm{NormECG}}
   =
   \left\{
   \mu_1,\mu_2,\ldots,\mu_C
   \right\},
\end{equation}
where each \(\mu_c\) is a time-warped representative of normal ECG segments within that cluster. Compared to naive pointwise averaging, these DTW-based barycenters better handle local temporal shifts and preserve waveform morphology more faithfully.

% \newpage
\section{Detailed Analysis of Extraction Module}
\label{app:extraction_module_analysis}

In this section, we illustrate how a slightly abnormal but still healthy ECG sample from a NormECG Cluster can be refined by Multi-dimensional DTW (MDTW) barycenter fusion. Specifically, we choose one central sample in the NormECG Cluster, as shown in Fig.~\ref{fig:selectedSample}, and show its corresponding aligned Codebook sample in Fig.~\ref{fig:MDTWSample}.

As illustrated, the originally selected central sample contained mild signal deviations in Leads~3 and~5, as indicated by the red box in Fig.~\ref{fig:selectedSample}, possibly due to acquisition noise or external factors. Although these deviations are not caused by true cardiovascular pathology, they could still trigger false alarms in a feature-based anomaly detection system. By contrast, after MDTW-based fusion with other similar normal samples, the resulting Codebook sample in Fig.~\ref{fig:MDTWSample} exhibits smoother and more typical waveforms in the affected leads, while retaining the overall morphology of the original signal in other leads. Consequently, this aligned Codebook better reflects the \textit{intrinsic} normal ECG pattern, facilitating more robust R-peak detection and providing a high-quality reference for subsequent anomaly-deviation calculations and reconstruction-based analyses.

\begin{figure}[!t]
\centering
\includegraphics[width=0.80\columnwidth]{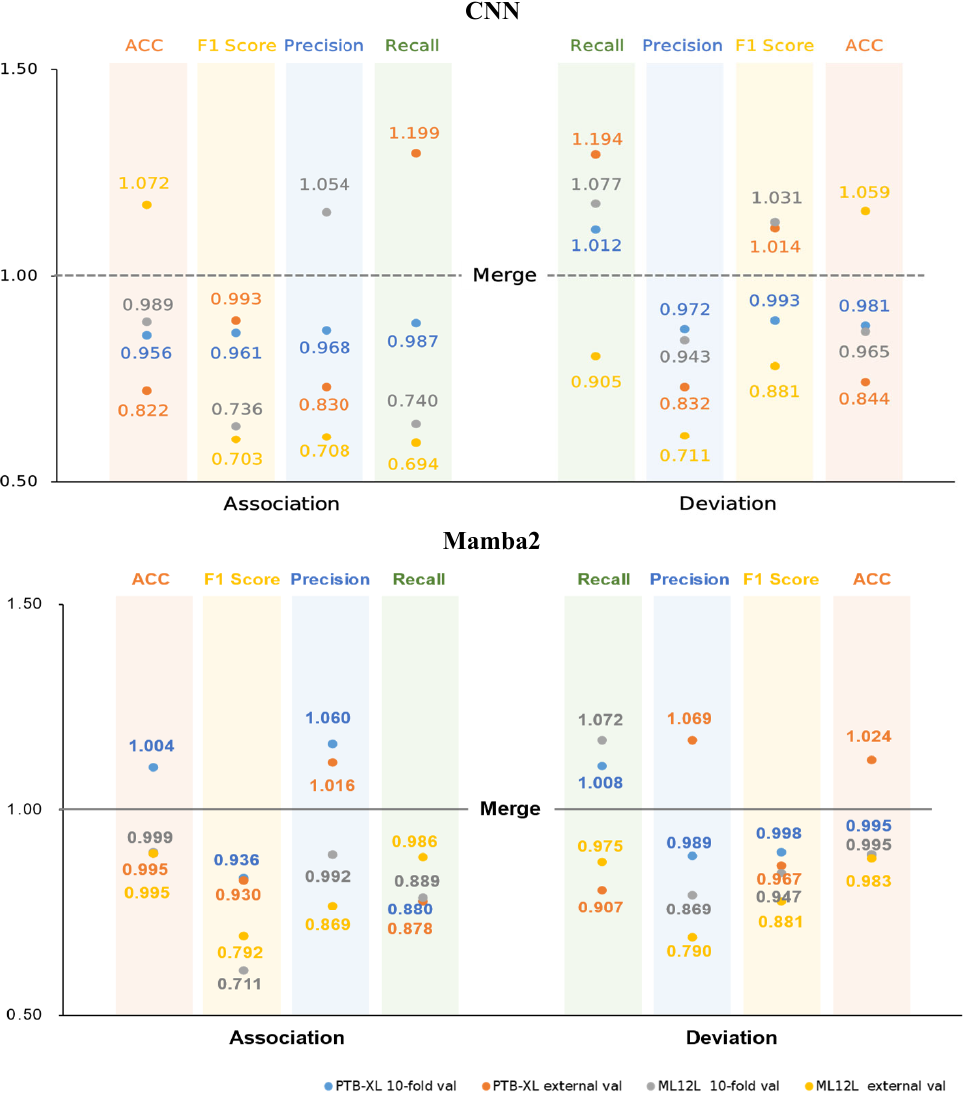}
\caption{Comparison of ablation performance using CNN and Mamba2 as classifiers. The left panel shows ARE, the right panel shows DLE, and the baseline denotes the combination of the two embeddings.}
\label{fig:ablation_embedding}
\end{figure}

\section{Embedding Component Contribution and Analysis}
\label{app:ablation_embedding}

In Fig.~\ref{fig:ablation_embedding}, the fused DLE+ARE embedding sets the baseline, i.e., 0\% change. We then measure the relative drop or gain when using only Association Reconstruction Embedding (ARE) or only Deviation Localization Embedding (DLE).

\subsection{ARE Only}

\begin{itemize}
  \item On \textbf{PTB-XL}, removing DLE causes Balanced Accuracy to fall by \(\sim 4\%\) and F1 by \(\sim 6\%\), since ARE emphasizes cross-lead correlations but misses localized amplitude deviations, e.g., minor ST shifts, that DLE captures. AUROC decreases by \(\sim 3\%\) overall, yet F1 for myocardial infarction improves by \(\sim 1.5\%\), reflecting ARE's strength in integrating multi-lead MI patterns.

  \item On \textbf{ML12L}, ARE-only drops Balanced Accuracy by \(\sim 5\%\) but boosts AUROC for conduction disorders by \(\sim 2\%\), because cross-lead attention highlights the broad QRS prolongation patterns there. F1 for repolarization abnormalities also gains \(\sim 1\%\), as ARE smooths out noisy local fluctuations.
\end{itemize}

\subsection{DLE Only}

\begin{itemize}
  \item On \textbf{PTB-XL}, removing ARE reduces F1 by \(\sim 7\%\) and AUROC by \(\sim 4\%\), while sensitivity actually increases by \(\sim 3\%\), showing that DLE's per-lead deviation maps are highly sensitive to subtle anomalies but lack ARE's inter-lead context to suppress false positives.

  \item On \textbf{ML12L}, DLE-only lowers overall Balanced Accuracy by \(\sim 6\%\), yet recall for hypertrophy and ischemia classes rises by \(\sim 4\%\), because DLE pinpoints local morphological deviations, e.g., voltage criteria, critical for these diagnoses.
\end{itemize}

\subsection{Detailed Interpretation}

\begin{itemize}
  \item \emph{DLE excels at detecting beat-level morphological anomalies}, e.g., ST/T deviations and voltage changes, and thus can increase sensitivity or recall for classes defined by sharp local deviations.

  \item \emph{ARE strengthens inter-lead consistency}, improving precision and class separability, as reflected in AUROC and F1 for broader conduction or MI patterns, but may smooth away isolated spikes that DLE would flag.

  \item \emph{Fused embedding leverages both}: DLE provides high recall for subtle, localized deviations, while ARE provides high precision by contextualizing anomalies across leads, yielding the best trade-off in Balanced Accuracy, F1, and AUROC.
\end{itemize}

\section{Parameter Settings and Model Structure}
\label{app:parameters}

In this section, we summarize the key hyperparameters and implementation choices used throughout our training pipeline.

\subsection{Model Architecture}

\begin{itemize}
  \item \(\texttt{d\_model}=256\): hidden dimension of the Mamba2 input and output.

  \item \(\texttt{d\_state}=64\): latent state dimension in the LTV-SSM.

  \item \(\texttt{d\_conv}=4\): kernel size for the small convolutional networks that parameterize the time-varying matrices \(A_t\), \(B_t\), \(C_t\), and \(D_t\).

  \item \(\texttt{dropout\_rate}=0.1\): AlphaDropout after the SELU activation.
\end{itemize}

\subsection{Optimization}

\begin{itemize}
  \item Loss: Smooth L1.

  \item Optimizer: Adam on all model and tokenizer parameters.

  \item Learning rate: \(\eta = 4 \times 10^{-5}\).

  \item Scheduler: \(\texttt{StepLR}(\texttt{step\_size}=5, \gamma=0.5)\), which decays \(\eta\) by half every 5 epochs.

  \item Number of epochs per split for pretraining: \(30\).

  \item Number of epochs in linear evaluation and generalization validation: \(20\).
\end{itemize}

\begin{table*}[!t]
\centering
\caption{Classification performance of different downstream classifiers using Baseline and DLE\&ARE as ECG representations on external clinical datasets. The best result in each column is shown in \textbf{bold}.}
\label{tab:observational_study_result}
\scriptsize
\setlength{\tabcolsep}{6pt}
\renewcommand{\arraystretch}{1.05}
\begin{tabular}{llccccc}
\toprule
\textbf{Classifier} & \textbf{Representation} 
& \textbf{Accuracy} & \textbf{F1 Score} & \textbf{Precision} & \textbf{Recall} & \textbf{AUC} \\
\midrule

\multirow{2}{*}{CNN}
& Baseline
& \(0.221\pm0.011\) & \(0.356\pm0.012\) & \(0.449\pm0.014\) & \(0.303\pm0.018\) & \(0.573\pm0.015\) \\
& DLE\&ARE
& \(\textbf{0.248}\pm\textbf{0.012}\) & \(\textbf{0.395}\pm\textbf{0.016}\) & \(\textbf{0.466}\pm\textbf{0.009}\) & \(\textbf{0.348}\pm\textbf{0.022}\) & \(\textbf{0.602}\pm\textbf{0.011}\) \\
\midrule

\multirow{2}{*}{Transformer}
& Baseline
& \(0.242\pm0.024\) & \(0.352\pm0.024\) & \(\textbf{0.532}\pm\textbf{0.009}\) & \(0.269\pm0.026\) & \(0.574\pm0.011\) \\
& DLE\&ARE
& \(\textbf{0.276}\pm\textbf{0.027}\) & \(\textbf{0.401}\pm\textbf{0.022}\) & \(0.512\pm0.011\) & \(\textbf{0.341}\pm\textbf{0.030}\) & \(\textbf{0.588}\pm\textbf{0.007}\) \\
\midrule

\multirow{2}{*}{Mamba}
& Baseline
& \(0.241\pm0.017\) & \(0.386\pm0.022\) & \(0.399\pm0.021\) & \(0.377\pm0.025\) & \(0.499\pm0.003\) \\
& DLE\&ARE
& \(\textbf{0.348}\pm\textbf{0.029}\) & \(\textbf{0.429}\pm\textbf{0.028}\) & \(\textbf{0.524}\pm\textbf{0.008}\) & \(\textbf{0.377}\pm\textbf{0.031}\) & \(\textbf{0.543}\pm\textbf{0.009}\) \\
\bottomrule
\end{tabular}
\end{table*}

These settings were chosen based on preliminary experiments balancing training stability, GPU memory constraints, all hyper-parameters remain the same for all models.

\subsection{Linear Evaluation Classifier Architectures}
\label{app:linear_eval_archs}

The architectures of the three linear evaluation classifiers used in our experiments are as follows:

\begin{itemize}
  \item \textbf{CNN Classifier:}
    \begin{itemize}
      \item Four sequential 1D convolutional blocks:
        \begin{itemize}
          \item Block 1: Conv1d \(36 \rightarrow 64\), kernel size \(7\), stride \(1\), padding \(3\); BatchNorm1d; ReLU; MaxPool1d with kernel size \(2\) and stride \(2\).
          \item Block 2: Conv1d \(64 \rightarrow 128\), kernel size \(5\), stride \(1\), padding \(2\); BatchNorm1d; ReLU; MaxPool1d with kernel size \(2\) and stride \(2\).
          \item Block 3: Conv1d \(128 \rightarrow 256\), kernel size \(3\), stride \(1\), padding \(1\); BatchNorm1d; ReLU; MaxPool1d with kernel size \(2\) and stride \(2\).
          \item Block 4: Conv1d \(256 \rightarrow 512\), kernel size \(3\), stride \(1\), padding \(1\); BatchNorm1d; ReLU; MaxPool1d with kernel size \(2\) and stride \(2\).
        \end{itemize}
      \item The output is flattened and passed through fully connected layers: \(512 \rightarrow 256 \rightarrow 64 \rightarrow\) number of classes.
    \end{itemize}

  \item \textbf{Transformer Classifier:}
    \begin{itemize}
      \item Transformer encoder: four layers, each with \(\texttt{d\_model}=720\), \(\texttt{nhead}=4\), feedforward dimension \(512\), and dropout rate \(0.1\).
      \item After encoding, the output is reshaped and flattened.
      \item Fully connected head: \(512 \rightarrow 256 \rightarrow 64 \rightarrow\) number of classes.
    \end{itemize}

  \item \textbf{Mamba Classifier:}
    \begin{itemize}
      \item Mamba block: \(\texttt{d\_model}=720\), \(\texttt{d\_state}=64\), and \(\texttt{d\_conv}=4\), followed by SELU activation and AlphaDropout with dropout rate \(0.1\).
      \item The output is reshaped to \((\mathrm{batch}, 36, 5000)\) and then flattened.
      \item Fully connected head: \(512 \rightarrow 256 \rightarrow 64 \rightarrow\) number of classes.
    \end{itemize}
\end{itemize}

% \begin{figure*}[!ht]
% \centering
% \caption{Performance comparison of different feature extractors across classifier architectures in real-world evaluation. }
% \label{fig:Observational}
% \end{figure*}

\section{SOTA Models Used for Performance Comparison}
\label{app:sota_models}

\subsection{SOTA Feature Extraction Models}

\textbf{DINOv2}~\cite{oquab2023dinov2} is a self-supervised vision transformer originally developed for masked image modeling. It is adapted here for ECG signals by treating them as images to leverage its strong unsupervised representation capabilities.

\textbf{TimesFM}~\cite{das2023decoder} is a decoder-only foundation model trained via contrastive learning on large-scale time series, capturing temporal dependencies well for sequence-level ECG modeling.

\textbf{Chronos}~\cite{ansari2024chronos} is a transformer-based model pretrained with time-warp augmentations and tokenized representations, designed to model complex temporal dynamics.

\textbf{MOIRAI}~\cite{woo2024unified} is a GAN-based anomaly detection model for multivariate physiological signals, capturing fine-grained correlations and supporting unsupervised ECG analysis.

\subsection{SOTA ECG-CVD Classification Models}

\textbf{ASTCL}~\cite{wang2023adversarial} is a transformer classifier trained with self-supervision for long ECG sequences, effective at learning discriminative features without labels.

\textbf{BeatGAN}~\cite{zhou2019beatgan} is a beat-level generative model that enables fine-grained rhythm and morphology analysis for arrhythmia detection.

\textbf{MERL}~\cite{liu2024zero} is a multi-resolution residual autoencoder for morphological anomaly detection using hierarchical representations and residual learning.

\textbf{ECGAD}~\cite{jiang2024self} is an autoencoder-based model that detects abnormal heartbeats via reconstruction error, tailored for ECG anomaly detection.

\section{Detailed Observational Study Results}
\label{app:observational_study_results}

We compared the characteristic performance of the baseline model and the fusion model on the real-world external clinical dataset generated by them. This comparison was conducted for three downstream classifier model architectures, including convolutional neural network, Transformer, and Mamba model. We evaluated two input types: baseline, i.e., original electrocardiogram, and the extracted feature combination of DLE + ARE. The performance was evaluated through accuracy (ACC), F1 score, precision, recall rate, and AUC. Table~\ref{tab:observational_study_result} provides detailed metric values. In most classifiers and evaluation metrics, the combined representation consistently outperformed other configurations, indicating its strong generalization ability in various real data scenarios.

\begin{table*}[!t]
\centering
\caption{Performance comparison of state-of-the-art self-supervised ECG classifiers on two unseen test sets, CPSC2020 and WGEX, evaluated using 10-fold cross-validation.}
\label{tab:sota-compare}
\scriptsize
\setlength{\tabcolsep}{4pt}
\renewcommand{\arraystretch}{1.05}
\begin{tabular*}{\textwidth}{@{\extracolsep{\fill}}lcccccccc}
\toprule
\textbf{Model} 
& \multicolumn{4}{c}{\textbf{CPSC2020}} 
& \multicolumn{4}{c}{\textbf{WGEX}} \\
\cmidrule(lr){2-5} \cmidrule(lr){6-9}
& \textbf{B-Acc} & \textbf{F1} & \textbf{AUC} & \textbf{Sensitivity}
& \textbf{B-Acc} & \textbf{F1} & \textbf{AUC} & \textbf{Sensitivity} \\
\midrule
ASTCL
& \(0.496\pm0.114\) & \(0.522\pm0.041\) & \(0.744\pm0.011\) & \(0.700\pm0.034\)
& \(0.530\pm0.057\) & \(0.462\pm0.070\) & \(0.724\pm0.096\) & \(0.699\pm0.111\) \\

BeatGAN
& \(0.480\pm0.069\) & \(0.546\pm0.027\) & \(0.762\pm0.088\) & \(0.690\pm0.006\)
& \(0.501\pm0.111\) & \(0.476\pm0.002\) & \(0.765\pm0.003\) & \(0.711\pm0.103\) \\

MERL
& \(0.243\pm0.077\) & \(0.462\pm0.103\) & \(0.724\pm0.083\) & \(0.686\pm0.097\)
& \(0.208\pm0.046\) & \(0.260\pm0.012\) & \(0.673\pm0.061\) & \(0.522\pm0.008\) \\

ECGAD
& \(0.502\pm0.062\) & \(0.436\pm0.042\) & \(0.740\pm0.107\) & \(0.723\pm0.030\)
& \(0.515\pm0.029\) & \(0.338\pm0.099\) & \(0.730\pm0.088\) & \(0.667\pm0.080\) \\

\textbf{Ours}
& \(\textbf{0.547}\pm\textbf{0.039}\) & \(\textbf{0.557}\pm\textbf{0.041}\) & \(\textbf{0.800}\pm\textbf{0.036}\) & \(\textbf{0.740}\pm\textbf{0.093}\)
& \(\textbf{0.584}\pm\textbf{0.049}\) & \(\textbf{0.538}\pm\textbf{0.015}\) & \(\textbf{0.809}\pm\textbf{0.061}\) & \(\textbf{0.712}\pm\textbf{0.092}\) \\
\bottomrule
\end{tabular*}
\end{table*}

\section{Full Table of Performance Comparison of SOTA Self-Supervised ECG Classifiers on the Unseen Test Sets CPSC2020 and WGEX}
\label{app:sota_comparison_full}

We present a comprehensive evaluation of state-of-the-art self-supervised ECG classification models on two external datasets, CPSC2020 and WGEX, both of which were not seen during training. As shown in Table~\ref{tab:sota-compare}, our method consistently outperforms existing approaches, including ASTCL, BeatGAN, MERL, and ECGAD, across all four evaluation metrics: balanced accuracy (B-Acc), F1 score, AUC, and sensitivity. Notably, our model achieves the highest AUC, 0.800 and 0.809, and balanced accuracy, 0.547 and 0.584, on both datasets, indicating superior generalization ability under distributional shifts. These results demonstrate that our representation learning framework yields more robust and discriminative features for unseen database scenarios.

\bibliographystyle{IEEEtran}
\bibliography{ref}